\documentclass[11pt]{article}

\usepackage[final]{acl}

\usepackage{times}
\usepackage{latexsym}
\usepackage{subcaption}  

\newcommand{\pmi}[2]{\ensuremath{#1_{\pm #2}}}
\newcommand{\bpmi}[2]{\ensuremath{\mathbf{#1}_{\pm #2}}}

\usepackage[T1]{fontenc}

\usepackage[utf8]{inputenc}
\usepackage{multirow} 
\usepackage{microtype}
\usepackage{inconsolata}

\usepackage{graphicx}
\usepackage{booktabs}
\usepackage{amsmath}
\usepackage{amssymb}
\usepackage{pifont}
\newcommand{\cmark}{\ding{51}} 
\newcommand{\xmark}{\ding{55}} 

\usepackage{xcolor}
\usepackage{listings}
\usepackage{tcolorbox}
\tcbuselibrary{breakable}

\lstdefinestyle{pywrap}{
  basicstyle=\ttfamily\footnotesize,
  breaklines=true,
  breakatwhitespace=true,
  columns=fullflexible,
  keepspaces=true,
  showstringspaces=false
}

\tcbset{
  codebox/.style={
    enhanced,
    breakable,
    sharp corners,
    boxrule=0.6pt,
    colframe=gray!75!black,
    colback=gray!3,
    left=1mm,right=1mm,top=0.5mm,bottom=0.5mm,
    before skip=6pt, after skip=6pt
  }
}
\usepackage{tcolorbox}
\tcbuselibrary{listings,breakable}

\usepackage{colortbl}

\newcommand{\pos}[1]{\textcolor{green!60!black}{#1}}
\newcommand{\nega}[1]{\textcolor{red!70!black}{#1}}

\title{CCTVBench: Contrastive Consistency Traffic VideoQA \\ Benchmark for Multimodal LLMs}

\author{
 \textbf{Xingcheng Zhou\textsuperscript{1}\thanks{Correspondence: \href{mailto:xingcheng.zhou@tum.de}{xingcheng.zhou@tum.de}}},
 \textbf{Hao Guo\textsuperscript{1}},
 \textbf{Rui Song\textsuperscript{2}},
 \textbf{Walter Zimmer\textsuperscript{2}},
 \textbf{Mingyu Liu\textsuperscript{1}}, \\
 \textbf{André Schamschurko\textsuperscript{1}},
 \textbf{Hu Cao\textsuperscript{1}},
 \textbf{Alois Knoll\textsuperscript{1}}
\\
\\
 \textsuperscript{1}Technical University of Munich 
 \textsuperscript{2}University of California, Los Angeles
}

\begin{document}
\maketitle

\begin{abstract}
Safety-critical traffic reasoning requires contrastive consistency: models must detect true hazards when an accident occurs, and reliably reject plausible-but-false hypotheses under near-identical counterfactual scenes. We present \textsc{CCTVBench}, a \textbf{C}ontrastive \textbf{C}onsistency \textbf{T}raffic \textbf{V}ideoQA \textbf{Bench}mark built on paired real accident videos and world-model-generated counterfactual counterparts, together with minimally different, mutually exclusive hypothesis questions. \textsc{CCTVBench} enforces a single structured decision pattern over each video question quadruple and provides actionable diagnostics that decompose failures into positive omission, positive swap, negative hallucination, and mutual-exclusivity violation, while separating video versus question consistency. Experiments across open-source and proprietary video LLMs reveal a large and persistent gap between standard per-instance QA metrics and quadruple-level contrastive consistency, with unreliable none-of-the-above rejection as a key bottleneck. Finally, we introduce C-TCD, a contrastive decoding approach leveraging a semantically exclusive counterpart video as the contrast input at inference time, improving both instance-level QA and contrastive consistency.
\end{abstract}
\section{Introduction}
\label{sec:introduction}

Vision–language models (VLMs) are increasingly applied to traffic scene understanding, roadway monitoring, and language-conditioned perception and decision making in autonomous driving systems~\cite{VisionLanguageModelsinAutonomousDrivin10531702,cui2025llm4adlargelanguagemodels}. However, current Video-LLMs are prone to hallucinating non-existent entities or events and to missing subtle but critical visual cues, while recent multi-modal studies therefore begin to systematically evaluate and mitigate such hallucinations~\cite{guan2023hallusionbench,leng2023mitigatingobjecthallucinationslargeVCD,zhang2025eventhallusiondiagnosingeventhallucinations}. In safety-critical traffic settings, this behavior manifests as false alarms and missed hazards that can propagate into unstable or unsafe downstream decisions. Beyond improving per-instance accuracy, it is therefore essential to ensure that models exhibit contrastive consistency: when the decisive outcome changes under tightly controlled scene edits, their predictions must adapt coherently to the underlying evidence. Concretely, models should behave consistently under scene-matched counterfactual edits, reject plausible-but-false alternatives, and support a calibrated none-of-the-above state instead of defaulting to overconfident but unsupported answers.

\begin{figure}[t]
  \centering
  \includegraphics[width=0.47\textwidth]{figure/TrafficBenchexample.jpg}
 \caption{Illustration of the contrastive quadruple evaluation protocol in CCTVBench. Given a contrastive video pair $(v^+, v^-)$ and a mutually exclusive question pair $(q^+, q^-)$, a consistent model should answer \text{Yes} only for $(v^+, q^+)$ and reject all other combinations.}  
 
 \label{fig:quadexample}
\end{figure}

At the same time, most existing traffic video QA benchmarks evaluate models on isolated video question pairs and report standard classification metrics \cite{qian2024nuscenesqa,zhou2025tumtrafficvideoqa,xu2021sutdtrafficqa,sima2024drivelm}, which provide limited signal on whether a model maintains a coherent decision rule when the decisive outcome changes while the surrounding scene context is largely preserved.  This leaves contrastive failures under scene-matched counterfactual edits underexplored. A model can achieve reasonable per-instance accuracy yet still answer \emph{Yes} on counterfactual negatives, endorse a plausible distractor on the same scene, or even accept mutually exclusive hypotheses, all unacceptable for risk-aware traffic decision making.
We therefore introduce \textsc{CCTVBench}, a \textbf{C}ontrastive \textbf{C}onsistency \textbf{T}raffic \textbf{V}ideoQA \textbf{Bench}mark.

\textsc{CCTVBench} is designed to make failures actionable.
Its metric suite measures video and question consistency, enforces a strict quadruple-level decision pattern, reports vision-language sensitivity, and decomposes failed quadruples into  positive omission, positive swap, negative hallucination, and mutual-exclusivity violation, as shown in Fig. \ref{fig:quadexample}.
This allows targeted diagnosis of whether errors stem from missed hazard detection on $v^+$, false positives on $v^-$, or confusion between paired hypotheses. The same counterfactual pairing also provides a stronger contrast signal at inference time. We therefore present C-TCD, which introduces the semantically exclusive counterpart video as a contrast input, showing that counterfactual pairs are not only for evaluation but also effective to improve consistency. Our contributions can be summarized as follows:

\begin{itemize}
    \item We introduce \textsc{CCTVBench}, a contrastive traffic VideoQA benchmark built on real anomaly footage, with contrastive video and question pairs across six categories.
    \item We propose a quadruple evaluation protocol with a diagnostic suite that measures cross-video and cross-question consistency and attributes quadruple failures to major patterns.
    \item We benchmark a broad set of video LLMs and reveal the gap between classification quality and contrastive consistency, highlighting robust rejection remains the main bottleneck.
    \item We propose C-TCD for inference-time contrastive decoding with semantically exclusive counterfactual pairs, demonstrating that counterfactual video pairs are not only for evaluation but also effective to improve model consistency.
\end{itemize}

\section{Related Work}

\paragraph{Traffic and Driving Vision-Language Benchmarks.}
VideoQA benchmarks in the traffic domain primarily study traffic event-centric reasoning and multi-agent interactions in real-world monitoring scenarios. SUTD-TrafficQA \citep{xu2021sutdtrafficqa} provides large-scale traffic video QA covering diverse traffic accidents, while TUMTraffic-VideoQA \citep{zhou2025tumtrafficvideoqa} studies incident-centric questions in roadside surveillance footage. In the driving domain, multiple works \citep{qian2024nuscenesqa,sima2024drivelm,marcu2024lingoqa} extend VQA to multi-view driving perception, targeting instruction-following and reasoning grounded in driving scenes understanding. Despite existing contrastive-pair benchmarks in general VQA \cite{li2025vidhalluc,qiu-etal-2024-valor,chandu2024certainlyuncertainbenchmarkmetric}, neither the traffic nor the driving VideoQA lines provide benchmarks explicitly designed for contrastive consistency evaluation, which is critical for VLM-based models.

\paragraph{Consistency and Hallucination Evaluation of Multimodal LLMs.}
Despite strong generation and instruction following, multimodal LLMs often hallucinate non-existent entities or events. In images, POPE~\citep{li-etal-2023-evaluating} proposes a polling-based protocol for stable hallucination measurement. Some benchmarks and mitigation work further motivate evidence-grounded evaluation \citealp{guan2023hallusionbench, lee-etal-2024-volcano}. In the video understanding domain, hallucinations are amplified by temporal ambiguity and long-range dependencies. VidHal \citep{choong2024vidhal} evaluates temporal hallucination via ordering-based probes, while VidHalluc \citep{li2025vidhalluc} targets temporally grounded hallucinations. Complementary to QA-only evaluation, contrastive setting formulations naturally support consistency checks \citep{liu2020violin,lei-etal-2020-likely,lei-etal-2020-tvqa, xiao2024trust}.

\paragraph{Controllable Driving World Models.}
World models for autonomous driving aim to simulate or forecast future observations under actions and interventions.
GAIA-2 \citep{russell2025gaia2controllablemultiviewgenerative} formulates driving world modeling as autoregressive sequence prediction conditioned on video, text, and actions. DrivingDiffusion \citep{li2024drivingdiffusion} synthesizes layout-guided multi-view driving videos using latent diffusion with cross-view and cross-frame consistency mechanisms. Drive-WM \citep{wang2024drivewm} generates controllable multi-view future videos and explores planning via imagined rollouts. MagicDrive \citep{gao2024magicdrive} provides diverse 3D geometry control for street-view generation and supports multi-view consistency. Recent world foundation model \citep{nvidia2025cosmosworldfoundationmodel, zheng2024opensorademocratizingefficientvideo} further scales controllable video world simulation and controllable video editing.

\section{\textsc{CCTVBench}}
\label{sec:benchmark}

In real-world driving, traffic accidents and anomalies are rare corner cases, and we empirically find that current world models struggle to synthesize physically plausible accident videos, whereas they can generate high-quality normal sequences. 


\subsection{Dataset Curation}
\label{sec:dataset-curation}

\begin{figure}[t]
  \centering
  \includegraphics[width=0.48\textwidth]{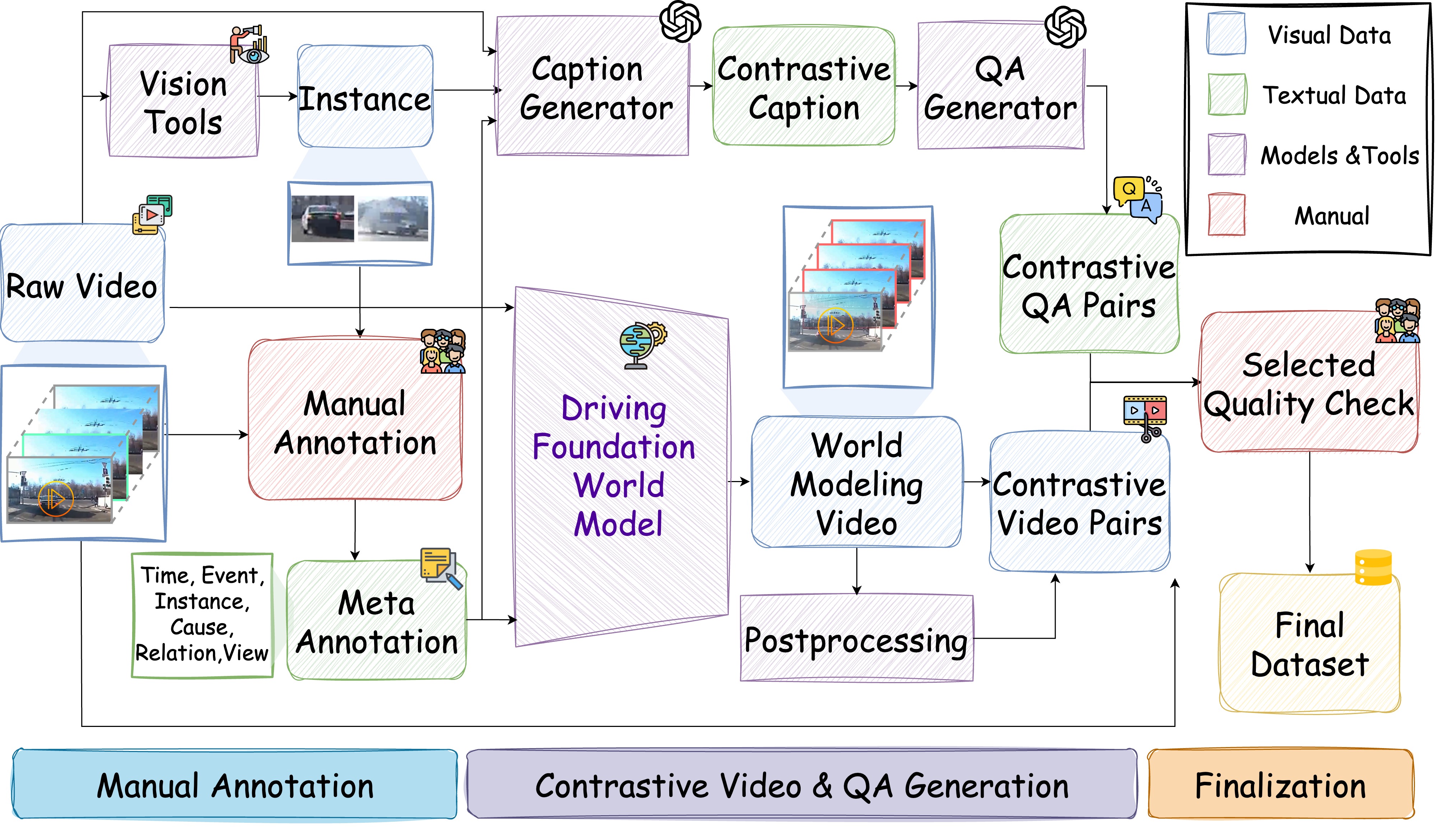}
 \caption{Dataset curation pipeline of CCTVBench.}
 \label{fig:flow}
\end{figure}

\begin{figure*}[t]
  \centering
  \includegraphics[width=\textwidth]{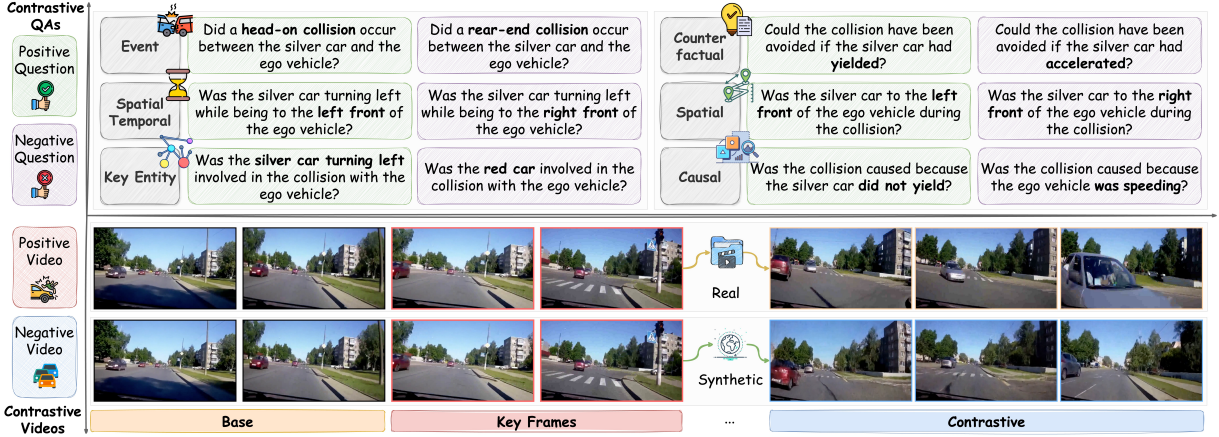}
  \caption{Quadruple Example of CCTVBench with contrastive video pair $(v_s^+,v_s^-)$ and corresponding contrastive question pairs $(q_{s,k}^+,q_{s,k}^-)$ across different categories.} 
  \label{fig:trafficbenchexample}
\end{figure*}

As illustrated in Fig.~\ref{fig:flow}, \textsc{CCTVBench} is curated to evaluate faithfulness via contrastive video pairs and mutually exclusive question pairs.
We start from open-source traffic and driving anomaly datasets and select representative accident scenes as positive videos $v_s^+$.
For each $v_s^+$, annotators segment the clip into three parts, as shown in Fig.~\ref{fig:trafficbenchexample}. A \emph{base} segment that does not affect the outcome, \emph{key frames} that determine the event trajectory, and a \emph{contrastive} segment targeted for counterfactual synthesis.
Besides, we run off-the-shelf detectors and trackers to extract traffic participants and coarse trajectories across the video. Annotators then refine instance identities and provide scene-level meta annotations, including involved instances, major causes, and critical spatio-temporal relations among key entities. To obtain minimally different yet semantically decisive video pairs, we generate the counterfactual variant $v_s^-$ using a driving world model ~\cite{nvidia2025cosmosworldfoundationmodel}.
We condition the generation on the shared base video prefix and synthesize the contrastive segment such that the accident does not occur, while preserving layout, illumination, and camera viewpoint as much as possible.
We then temporally align $(v_s^+, v_s^-)$ and apply quality filtering to remove videos with obvious artifacts, viewpoint drift, and key-entity inconsistencies, ensuring $v_s^-$ remains a valid negative control for the target hypotheses.

Given the aligned video pair and the human meta annotations, we adopt GPT-5 to generate paired contrastive questions across multiple categories, spanning event, key entity, temporal, spatial, spatio-temporal, causal, and counterfactual. We further constrain $q_{s,k}^-$ to be a plausible but verifiably false distractor on $v_s^+$. By design, $(q_{s,k}^+, q_{s,k}^-)$ are mutually exclusive but not collectively exhaustive: they cannot both be true, and on the negative-control video $v_s^-$ we expect both to be false with respect to the target hypotheses. By construction, each QA pair follows a fixed four-way decision pattern: on the positive video $v_s^+$, the positive question $q_{s,k}^+$ should be answered \emph{Yes} while its counterpart $q_{s,k}^-$ should be \emph{No}; on the counterfactual negative-control video $v_s^-$, \emph{both} questions should be answered \emph{No}. Finally, we conduct human validation and remove samples with generation artifacts, semantic mismatch, viewpoint drift, key-entity inconsistency, or ambiguous question grounding, yielding the finalized dataset. This review ensures that each $(v^+, v^-)$ pair remains a valid contrastive control and that each question pair is visually grounded and mutually exclusive.

\subsection{Dataset Overview}
\label{sec:dataset-stats}

\textsc{CCTVBench} is a contrastive traffic video QA benchmark with 305 contrastive scenes, 305 positive–negative video pairs $v^+$ and $v^-$, and 1,776 mutually exclusive question pairs, yielding 7,104 binary video question instances across six categories: Event, Key-Entity, Spatial, Spatial-Temporal, Causal, and Counterfactual.

\begin{figure}[h]
  \centering
  \begin{minipage}{0.28\textwidth}
    \centering
    \includegraphics[width=\textwidth]{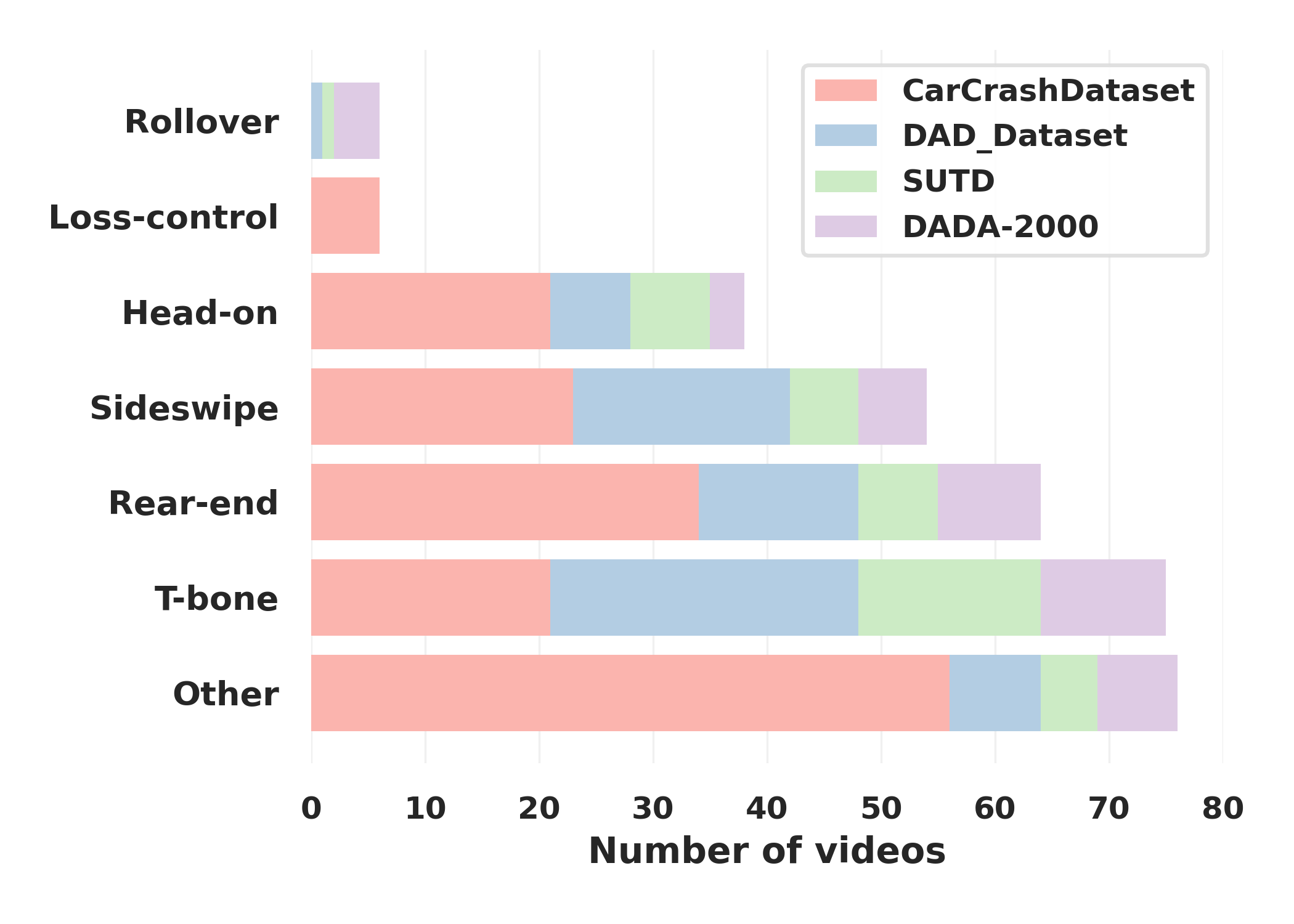}
  \end{minipage}
  \hfill
  \begin{minipage}{0.19\textwidth}
    \centering
    \includegraphics[width=\textwidth]{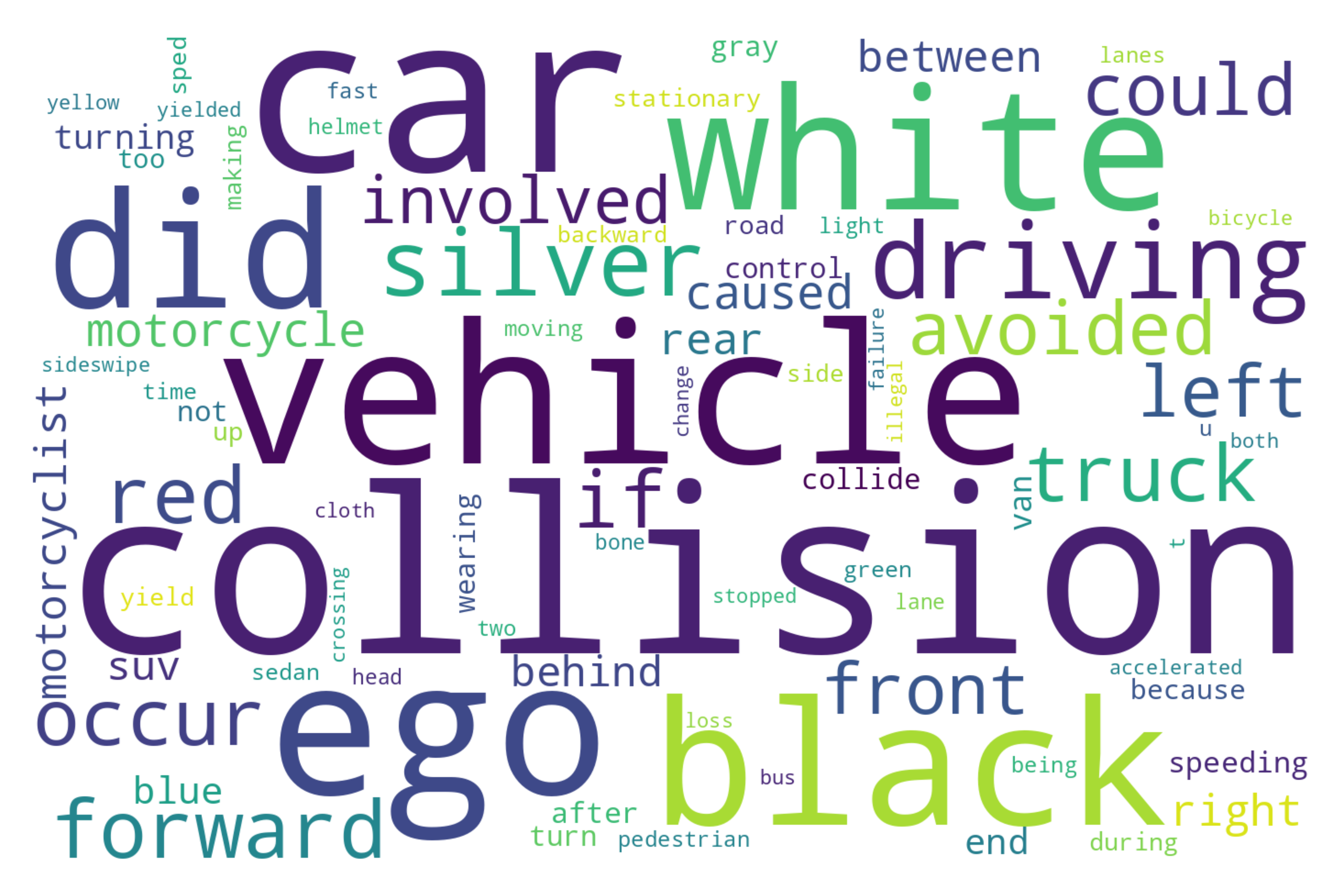}
  \end{minipage}
  \caption{Left: Distribution of event types from various sources. Right: Word cloud of event scene caption.}
  \label{fig:dataset_stats}
\end{figure}

Fig.~\ref{fig:dataset_stats} (left) summarizes the distribution of high-level crash types such as rear-end, T-bone, and sideswipe across various accident video datasets \cite{DADA-2000,DADDataset,CCDDataset}, covering a broad range of recording sources. Fig.~\ref{fig:dataset_stats} (right) shows the question vocabulary, dominated by accident-centric words such as car, vehicle, and collision.  

\subsection{Counterfactual Quality and Reliability}

We quantify generator-induced bias and distribution shift in counterfactual videos using two complementary analyses: artifact leakage and distributional similarity.

\paragraph{Artifact Leakage.} Using frozen R3D-18 features and a linear classifier, distinguishing counterfactual from real videos yields near-chance performance (AUC 56.7\%, accuracy 53.6\%), indicating that counterfactual videos do not exhibit strong generator-specific artifacts that could be exploited as shortcut signals.

\paragraph{Distributional Similarity.} We measure FVD and FID between counterfactual and real videos. Counterfactual videos are substantially closer to real accident videos (FVD 22.44, FID 32.63) than to unrelated normal driving footage (FVD 35.73, FID 105.00), suggesting a moderate distribution shift.

Overall, these results indicate that generator-induced bias is limited and that counterfactual videos remain reasonably aligned with the real data distribution.

\subsection{Contrastive Consistency Evaluation}
\label{sec:contrastive_evaluation}

The core of \textsc{CCTVBench} is a contrastive quadruple structure that evaluates whether a model can maintain a coherent decision pattern across contrastive paired videos and mutually exclusive questions. Let $K_s$ be the number of question pairs for scene $s$, $N_{\text{pairs}} = \sum_s K_s$ the total number of $(s,k)$ pairs, and $\sum_{s,k}$ denote summation over all such pairs. For brevity, we denote the binary outputs


\begin{equation}
\resizebox{0.7\linewidth}{!}{$
\begin{aligned}
\hat{y}^{++}_{s,k} &= \hat{y}(v_s^+, q_{s,k}^+), &
\hat{y}^{+-}_{s,k} &= \hat{y}(v_s^+, q_{s,k}^-), \\
\hat{y}^{-+}_{s,k} &= \hat{y}(v_s^-, q_{s,k}^+), &
\hat{y}^{--}_{s,k} &= \hat{y}(v_s^-, q_{s,k}^-)
\end{aligned}
$}
\label{eq:pair}
\end{equation}

with $\hat{y}(\cdot)\in\{0,1\}$ denoting the final binary prediction. 

\paragraph{Quadruple Correctness.}

The benchmark is constructed such that each contrastive quadruple has a single consistent semantic pattern: $q_{s,k}^+$ is true only on $v_s^+$, $q_{s,k}^-$ is false on both videos, and $v_s^-$ is a genuine \textit{none-of-the-above} counterfactual. We count a pair $(s,k)$ as fully consistent only if the model recovers the entire pattern:


\begin{equation}
\resizebox{\linewidth}{!}{$
\begin{aligned}
\mathrm{QuadCr}(\cdot)
&=
\mathbf{1}\!\left[
\hat{y}^{++}=1 \wedge
\hat{y}^{+-}=0 \wedge
\hat{y}^{-+}=0 \wedge
\hat{y}^{--}=0
\right], \\
\mathrm{QuadAcc}
&=
\frac{1}{N_{\text{pairs}}}
\sum_{s,k} \mathrm{QuadCr}(s,k)
\end{aligned}
$}
\label{eq:quadacc}
\end{equation} 

$\mathrm{QuadAcc}$ therefore measures how often a model simultaneously detects the target event in the positive video, rejects the mutually exclusive alternative on the same scene, and assigns negative answers to both hypotheses on the counterfactual video. Unlike single-video accuracy, it only gives credit when the entire contrastive VQA pair is correct, making it the strictest consistency metric in our benchmark.

\paragraph{Video Consistency.}
Video consistency evaluates whether model predictions genuinely respond to changes in the video when question is kept fixed. We therefore consider two complementary quantities. For positive query $q^+$, the desired behavior is a clean transition from \textit{Yes} on the positive video $v_s^+$ to \textit{No} on the counterfactual $v_s^-$:

\begin{equation}
\resizebox{0.85\linewidth}{!}{$
\mathrm{Contr@q^+}
=
\frac{1}{N_{\text{pairs}}}
\sum_{s,k}
\mathbf{1}\big[
\hat{y}^{++}_{s,k}=1 \wedge \hat{y}^{-+}_{s,k}=0
\big]
$}
\label{eq:contr_qplus_main}
\end{equation}

For the mutually exclusive alternative query $q^-$, the model should consistently reject the hypothesis on both videos:

\begin{equation}
\resizebox{0.85\linewidth}{!}{$
\mathrm{Reject@q^-}
=
\frac{1}{N_{\text{pairs}}}
\sum_{s,k}
\mathbf{1}\big[
\hat{y}^{+-}_{s,k}=0 \wedge \hat{y}^{--}_{s,k}=0
\big]
$}
\label{eq:reject_qminus_main}
\end{equation}

In the tables, \emph{Video Consistency} is reported as the average of Contr@q$^+$ and Reject@q$^-$, summarizing how reliably a model exploits the contrast between $v_s^+$ and $v_s^-$ under fixed queries, both required by the quadruple pattern.

\paragraph{Question Consistency.}
Question consistency, in turn, probes whether the model separates the two \emph{linguistic} hypotheses $q_{s,k}^+$ and $q_{s,k}^-$ when the video is fixed.  
On the positive video $v_s^+$, the desired pattern is to endorse the target proposition and reject its mutually exclusive alternative:

\begin{equation}
\resizebox{0.85\linewidth}{!}{$
\mathrm{Contr@v^+}
=
\frac{1}{N_{\text{pairs}}}
\sum_{s,k}
\mathbf{1}\big[
\hat{y}^{++}_{s,k}=1 \wedge \hat{y}^{+-}_{s,k}=0
\big]
$}
\label{eq:contr_vplus_main}
\end{equation}

On the counterfactual video $v_s^-$, both formulations should be rejected:

\begin{equation}
\resizebox{0.85\linewidth}{!}{$
\mathrm{Reject@v^-}
=
\frac{1}{N_{\text{pairs}}}
\sum_{s,k}
\mathbf{1}\big[
\hat{y}^{-+}_{s,k}=0 \wedge \hat{y}^{--}_{s,k}=0
\big]
$}
\label{eq:reject_vminus_main}
\end{equation}

We report \emph{Question Consistency} as the average of Contr@v$^+$ and Reject@v$^-$, balancing mutual-exclusivity handling on $v^+$ and none-of-the-above rejection on $v^-$.

\paragraph{Global Classification Metrics.}
To relate this contrastive protocol to standard QA performance, we additionally report balanced accuracy (BaAcc) and a squashed Matthews correlation coefficient
\begin{equation}
\resizebox{0.55\linewidth}{!}{$
\mathrm{MCCScore}
=
\left(\frac{\mathrm{MCC}+1}{2}\right)^2
$}
\label{eq:mccscore_main}
\end{equation}
computed over all VQA instances. MCCScore follows the setting of $Score_{cls}$ in \cite{li2025vidhalluc}. Both summarize plain binary classification quality on our data, while the contrastive metrics above reveal how that quality decomposes when video and question are systematically perturbed.


\subsection{Behavior Diagnosis}
\label{sec:hall_vl_diag}

Contrastive consistency scores indicate whether a model succeeds on a quadruple, but not how it fails. To analyze typical error patterns in traffic reasoning, we introduce two diagnostic families tailored to our setting: failure-mode decomposition over failed quadruples, and vision-language sensitivity within the same quadruple structure.

\paragraph{Failure Mode Decomposition.}
\label{sec:failure_modes}


To expose the typical failure reasons of existing Video LLMs, we focus on the subset of failed question pairs $F$ and decompose them into 4 major failure modes: positive omission, positive swap, negative hallucination, and mutual-exclusivity violation. For each failure mode, we report its failure rate as the proportion of $(s,k)\in F$ that satisfy the corresponding condition. Positive omission measures how often the model misses the target hypothesis on positive video $v_s^+$:

\begin{equation}
\resizebox{0.65\linewidth}{!}{$
\mathrm{PosOmiss}
=
\frac{1}{|F|}
\sum_{}
\mathbf{1}\!\left[\hat{y}^{++}=0\right]
$}
\end{equation}

Positive swap measures how often the model answers \emph{Yes} to $q_{s,k}^-$ on positive video $v_s^+$:
\begin{equation}
\resizebox{0.65\linewidth}{!}{$
\mathrm{PosSwap}
=
\frac{1}{|F|}
\sum_{}
\mathbf{1}\!\left[\hat{y}^{+-}=1\right]
$}
\end{equation}

Negative hallucination quantifies violations of the none-of-the-above requirement on $v_s^-$:
\begin{equation}
\resizebox{0.75\linewidth}{!}{$
\mathrm{NegHall}
=
\frac{1}{|F|}
\sum_{}
\mathbf{1}\!\left[\hat{y}^{-+}=1 \,\vee\, \hat{y}^{--}=1\right]
$}
\end{equation}

Mutual-exclusivity violation measures how often the model simultaneously breaks mutually exclusive hypotheses on the same video, both on $v_s^+$ and $v_s^-$:

\begin{equation}
\resizebox{0.95\linewidth}{!}{$
\mathrm{MEViol}
=
\frac{1}{|F|}
\sum_{}
\mathbf{1}\!\Big[
(\hat{y}^{++}=1 \wedge \hat{y}^{+-}=1)
\;\vee\;
(\hat{y}^{-+}=1 \wedge \hat{y}^{--}=1)
\Big]
$}
\end{equation}
Note that a failed pair can satisfy multiple reasons, so the 4 failure rates are not mutually exclusive and do not necessarily sum to 1.
 
\paragraph{Vision-Language Sensitivity.}
Vision-language sensitivity metrics show how strongly model predictions depend on visual evidence or textual modality. Video sensitivity (VS) measures the effect of swapping $v_s^+$ and $v_s^-$ under the positive query:

\begin{equation}
\resizebox{0.6 \linewidth}{!}{$
\mathrm{VS}
=
\frac{1}{N_{\text{pairs}}}
\sum_{s,k}
\big|\hat{y}^{++}_{s,k} - \hat{y}^{-+}_{s,k}\big|
$}
\label{eq:vs_main}
\end{equation}
while question sensitivity (QS) measures the effect of swapping $q_{s,k}^+$ and $q_{s,k}^-$ on the positive video:
\begin{equation}
\resizebox{0.6 \linewidth}{!}{$
\mathrm{QS}
=
\frac{1}{N_{\text{pairs}}}
\sum_{s,k}
\big|\hat{y}^{++}_{s,k} - \hat{y}^{+-}_{s,k}\big|
$}
\label{eq:qs_main}
\end{equation}

We aggregate them into a visual reliance index and a global vision-language balance score:
\begin{equation}
\resizebox{0.8 \linewidth}{!}{$
\mathrm{VRI} = \frac{\mathrm{VS}}{\mathrm{VS}+\mathrm{QS}}, \quad
\mathrm{GVRS} = 2 \cdot \mathrm{VRI} \cdot \mathrm{QS}
$}
\label{eq:vri_gvrs}
\end{equation}

VRI quantifies the relative reliance on the video modality, whereas GVRS captures joint sensitivity to both visual changes and question wording. It attains high values only when the model reacts strongly to perturbations in both the vision and language branches.

\begin{equation}
\resizebox{0.89 \linewidth}{!}{$
\mathrm{SVE}
=
\frac{1}{N_{\text{pairs}}}
\sum_{s,k}
\mathbf{1}\Big[
(\hat{y}^{++}_{s,k} \neq \hat{y}^{-+}_{s,k})
\wedge
(\hat{y}^{+-}_{s,k} = \hat{y}^{--}_{s,k})
\Big]
$}
\label{eq:sve_main}
\end{equation}

To capture more selective visual grounding, we additionally report the selective video effect (SVE), which counts cases where changing the video flips the answer only for the positive query $q^+$ while leaving responses to $q^-$ unchanged.

\section{Experiments}
\label{sec:experiments}

\begin{table*}[t]
\centering
\small
\setlength{\tabcolsep}{4pt}
\caption{Quadruple-consistency evaluation on \textsc{CCTVBench}. All values are percentages. Each entry is reported as mean with confidence interval, where the interval is the larger absolute deviation, in percentage points, between the mean and the upper bound of the 95\% bootstrap confidence interval.}
\label{tab:quadacc_detailed}
\resizebox{\textwidth}{!}{%
\begin{tabular}{l c ccc ccc ccc}
\toprule
\multirow{2}{*}{\textbf{Model}} & \multirow{2}{*}{\textbf{Size}} &
\multicolumn{3}{c}{\textbf{Video Consistency}} &
\multicolumn{3}{c}{\textbf{Question Consistency}} &
\multirow{2}{*}{\textbf{BaAcc}$\uparrow$} &
\multirow{2}{*}{\textbf{MCCScore}$\uparrow$} &
\multirow{2}{*}{\textbf{QuadAcc}$\uparrow$} \\
\cmidrule(lr){3-5}\cmidrule(lr){6-8}
& &
\textbf{Contr@q+}$\uparrow$ & \textbf{Reject@q-}$\uparrow$ & \textbf{Overall}$\uparrow$ &
\textbf{Contr@v+}$\uparrow$ & \textbf{Reject@v-}$\uparrow$ & \textbf{Overall}$\uparrow$ &
& & \\
\midrule
\multicolumn{11}{c}{\textit{Open-source Models}} \\
\midrule
InternVL3.5 & 2B & \pmi{13.05}{1.25} & \pmi{63.79}{1.76} & \pmi{38.42}{1.06} & \pmi{24.23}{1.57} & \pmi{53.54}{1.77} & \pmi{38.88}{1.01} & \pmi{58.46}{0.97} & \pmi{33.17}{1.02} & \pmi{7.05}{0.98} \\
InternVL3.5 & 8B & \pmi{20.88}{1.45} & \pmi{64.92}{1.77} & \pmi{42.90}{1.06} & \bpmi{38.58}{1.81} & \pmi{48.49}{1.83} & \pmi{43.53}{1.09} & \cellcolor{black!15}\bpmi{64.85}{1.10} & \cellcolor{black!15}\bpmi{39.89}{1.24} & \cellcolor{black!15}\bpmi{11.38}{1.16} \\

PLLaVA & 7B & \pmi{8.89}{1.04} & \pmi{80.75}{1.45} & \pmi{44.82}{0.74} & \pmi{22.35}{1.47} & \pmi{68.95}{1.62} & \pmi{45.65}{0.74} & \pmi{57.38}{0.90} & \pmi{33.35}{1.04} & \pmi{3.90}{0.71} \\
PLLaVA & 13B & \pmi{13.87}{1.26} & \pmi{66.67}{1.68} & \pmi{40.27}{0.95} & \pmi{29.06}{1.58} & \pmi{52.39}{1.84} & \pmi{40.73}{0.94} & \pmi{59.81}{1.09} & \pmi{34.70}{1.14} & \pmi{6.19}{0.90} \\

Qwen3-VL & 2B & \pmi{17.02}{1.37} & \pmi{63.51}{1.77} & \pmi{40.26}{1.00} & \pmi{34.21}{1.70} & \pmi{47.94}{1.76} & \pmi{41.07}{0.99} & \pmi{62.96}{1.01} & \pmi{37.77}{1.08} & \pmi{7.62}{0.98} \\
Qwen3-VL & 8B & \cellcolor{black!15}\pmi{13.01}{1.26} & \cellcolor{black!15}\bpmi{91.05}{1.03} & \cellcolor{black!15}\bpmi{52.03}{0.75} & \cellcolor{black!15}\pmi{27.85}{1.65} & \cellcolor{black!15}\bpmi{76.38}{1.54} & \cellcolor{black!15}\bpmi{52.11}{0.74} & \pmi{60.26}{1.00} & \pmi{38.62}{1.26} & \pmi{10.65}{1.13} \\

VideoLLaMA3 & 2B & \bpmi{21.65}{1.40} & \pmi{31.16}{1.66} & \pmi{26.41}{1.11} & \pmi{27.44}{1.62} & \pmi{28.28}{1.63} & \pmi{27.86}{1.13} & \pmi{55.38}{1.17} & \pmi{29.89}{1.10} & \pmi{6.48}{0.88} \\
VideoLLaMA3 & 7B & \pmi{20.04}{1.48} & \pmi{36.80}{1.78} & \pmi{28.42}{1.12} & \pmi{36.41}{1.84} & \pmi{21.81}{1.49} & \pmi{29.11}{1.13} & \pmi{58.47}{1.26} & \pmi{32.91}{1.24} & \pmi{6.58}{0.92} \\
VideoChatGPT & 7B & \pmi{16.21}{1.32} & \pmi{21.45}{1.50} & \pmi{18.83}{1.01} & \pmi{18.59}{1.50} & \pmi{19.81}{1.41} & \pmi{19.20}{1.01} & \pmi{49.85}{1.14} & \pmi{24.87}{1.02} & \pmi{4.13}{0.71} \\
Phi-4-Multimodal & 14B & \pmi{11.23}{1.16} & \pmi{40.07}{1.83} & \pmi{25.65}{1.01} & \pmi{27.26}{1.67} & \pmi{25.38}{1.52} & \pmi{26.32}{1.02} & \pmi{57.07}{1.09} & \pmi{31.61}{1.07} & \pmi{2.92}{0.60} \\

\midrule
\multicolumn{11}{c}{\textit{Proprietary Models}} \\
\midrule
Gemini-2.5 & flash & \pmi{24.48}{1.48} & \pmi{83.79}{1.32} & \pmi{54.13}{0.99} & \pmi{55.56}{1.75} & \pmi{48.91}{1.75} & \pmi{52.24}{1.10} & \pmi{69.66}{1.18} & \pmi{46.43}{1.45} & \pmi{20.87}{1.39} \\
Gemini-2.5 & flash-lite & \pmi{21.43}{1.46} & \pmi{86.95}{1.25} & \pmi{54.19}{0.94} & \bpmi{59.42}{1.79} & \pmi{48.86}{1.86} & \pmi{54.14}{0.93} & \pmi{70.37}{1.16} & \pmi{47.71}{1.39} & \pmi{18.28}{1.39} \\
GPT-5 & nano & \pmi{13.92}{1.27} & \bpmi{91.19}{1.03} & \pmi{52.55}{0.81} & \pmi{26.24}{1.58} & \bpmi{68.39}{1.66} & \pmi{47.31}{1.00} & \pmi{57.46}{1.04} & \pmi{34.43}{1.35} & \pmi{12.60}{1.18} \\
GPT-5 & mini & \cellcolor{black!15}\bpmi{29.07}{1.66} & \cellcolor{black!15}\pmi{87.13}{1.26} & \cellcolor{black!15}\bpmi{58.10}{1.04} & \cellcolor{black!15}\pmi{56.64}{1.88} & \cellcolor{black!15}\pmi{55.47}{1.76} & \cellcolor{black!15}\bpmi{56.05}{1.08} & \cellcolor{black!15}\bpmi{70.43}{1.23} & \cellcolor{black!15}\bpmi{48.43}{1.56} & \cellcolor{black!15}\bpmi{25.85}{1.57} \\

\bottomrule
\end{tabular}%
}
\end{table*}

\subsection{Models}
We evaluate a diverse set of video LLMs spanning multiple well-known VLMs. Our open-source lineup includes InternVL3.5 \cite{wang2025internvl35advancingopensourcemultimodal}, Qwen2.5-VL \cite{bai2025qwen25vltechnicalreport}, Qwen3-VL \cite{bai2025qwen3vltechnicalreport}, PLLaVA \cite{xu2024pllavaparameterfreellava}, VideoLLaMA3 \cite{zhang2025videollama3frontiermultimodal}, VideoChatGPT \cite{maaz2024videochatgptdetailedvideounderstanding}, and phi-4-multimodal \cite{microsoft2025phi4minitechnicalreportcompact}. We also include proprietary API-based models from Gemini-2.5 \cite{comanici2025gemini25pushingfrontier} and GPT-5 \cite{openai2024gpt4technicalreport}, i.e., Gemini-2.5 flash/flash-lite; GPT-5 mini/nano. 


\subsection{Experimental Setup}
\label{sec:setup}

All open-sourced models are evaluated under a unified binary QA interface. Each query is formatted with the instruction: \textit{You are given a traffic video and a question. Answer based only on the video with exactly one word: Yes or No. Question: \{QUESTION\}.} We instruct the model to answer with a single word (\texttt{Yes} or \texttt{No}) and map the output to a binary label. We use the default model decoding strategy with a temperature as 0. For proprietary models, we run the model in a video-level batch to reduce the cost of closed-source API evaluations, while then parsing each QA as an independent binary response.

\subsection{Contrastive Decoding Inference}
\label{sec:cd_evaluation}

Recent work has shown that training-free contrastive decoding can mitigate hallucinations in large VLMs by contrasting the output distributions of an original input with those of a deliberately degraded counterpart during inference. Therefore, we follow VCD~\cite{leng2023mitigatingobjecthallucinationslargeVCD} and TCD~\cite{zhang2025eventhallusiondiagnosingeventhallucinations} to examine how these methods behave under our contrastive evaluation protocol. Building on the paired counterfactual videos provided by our benchmark, we additionally present contrastive temporal contrastive decoding (C-TCD). Instead of constructing the contrast input by generic visual corruption or temporal degradation, C-TCD uses the semantically exclusive counterpart video from the same scene as the contrast signal. At each decoding step $t$ for a query $(v,q)$, we obtain logits under the original video $z^{\text{ori}}_t$ and under a contrastive video $z^{\text{con}}_t$, and fuse them as

\begin{equation}
\resizebox{0.45\linewidth}{!}{$
z_t
=
(1+\alpha)\, z^{\text{ori}}_t
-
\alpha\, z^{\text{con}}_t,
$}
\label{eq:cd_fusion}
\end{equation}
where $\alpha\ge 0$ controls the contrast strength and $\alpha=0$ reduces to vanilla decoding.

\begin{figure}[t]
  \centering
  \includegraphics[width=0.48\textwidth]{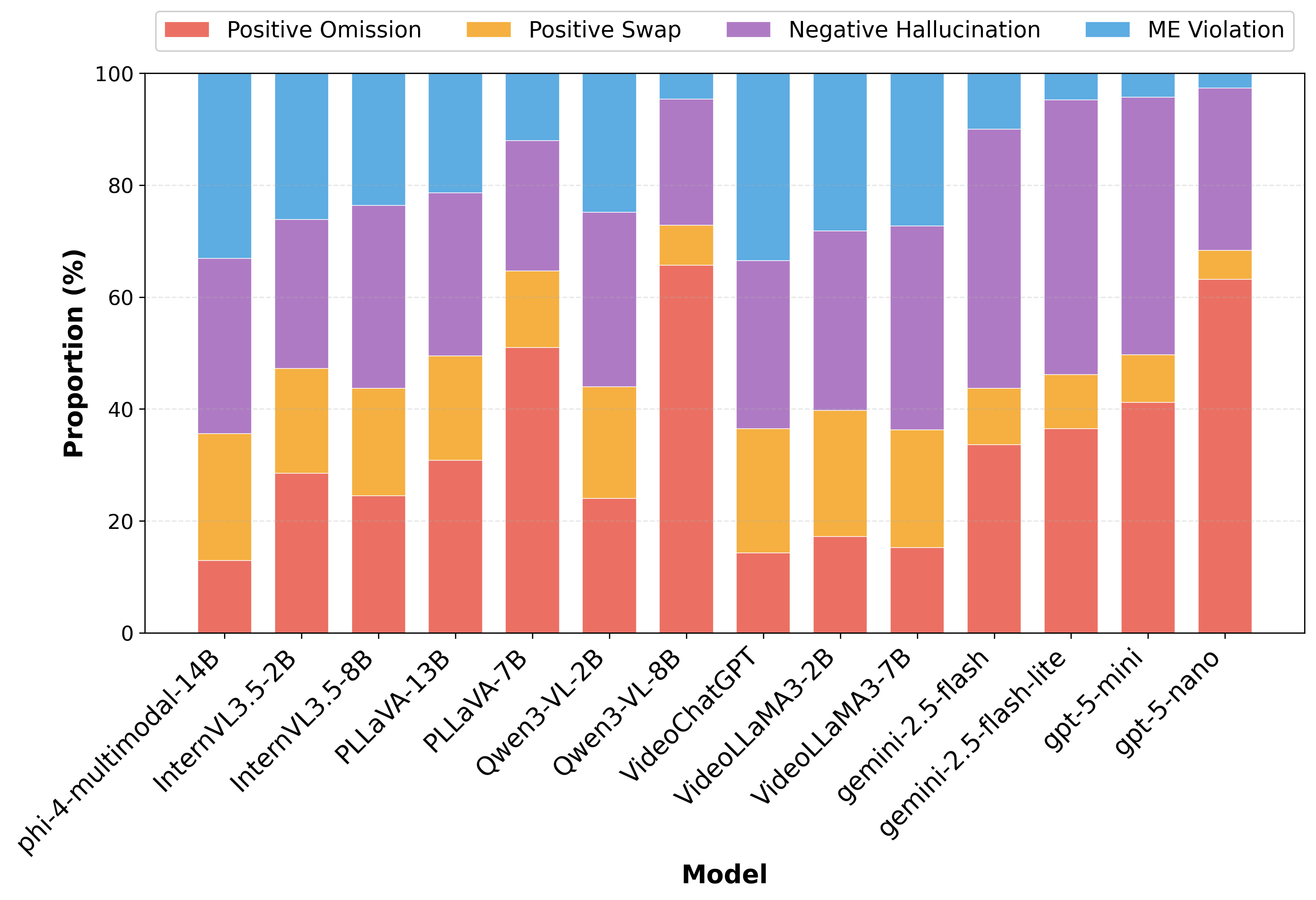}
  \caption{Normalized four failure-mode composition across various models. }
  \label{fig:failure_mode_stacked}
\end{figure}

The three variants differ in how $z^{\text{con}}_t$ is constructed. VCD and TCD rely on heuristic degradations of the same input video. In VCD, $z^{\text{con}}_t$ is computed on a visually corrupted video $v^{-}_{\text{vcd}}$ obtained by injecting diffusion noise into frames, which preserves coarse layout but removes fine-grained evidence. In TCD, $z^{\text{con}}_t$ is computed on a temporally degraded video $v^{-}_{\text{tcd}}$ produced by strong temporal downsampling, which blurs temporal cues while keeping appearance largely intact. 

In contrast, C-TCD uses a semantically decisive, scene-matched counterpart video provided by \textsc{CCTVBench}. We compute $z^{\text{con}}_t$ on the paired counterfactual video from the same scene, denoted $v^{-}_{\text{c-tcd}}$: for a positive video $v_s^+$ we use its paired negative $v_s^-$, and for $v_s^-$ we use $v_s^+$. C-TCD therefore replaces generic corruption or temporal degradation with an aligned counterfactual contrast signal, which directly matches the benchmark design and provides a more targeted semantic regularizer at inference time.

\paragraph{Assumptions and interpretation.}
C-TCD is a training-free inference-time calibration method. It does not modify model parameters or improve intrinsic visual representations, but instead adjusts prediction confidence using semantically matched counterfactual evidence. It does not assume that the counterfactual video perfectly replicates the original scene with only the target signal removed. Instead, it relies on a weaker condition: the counterfactual reduces causal evidence for the positive hypothesis while preserving most background context. Under this setting, the logit difference between the original and counterfactual inputs acts as a contrastive calibration signal. 

\begin{figure*}[t]
  \centering

  \begin{subfigure}[t]{0.31\textwidth}
    \centering
    \includegraphics[width=\textwidth]{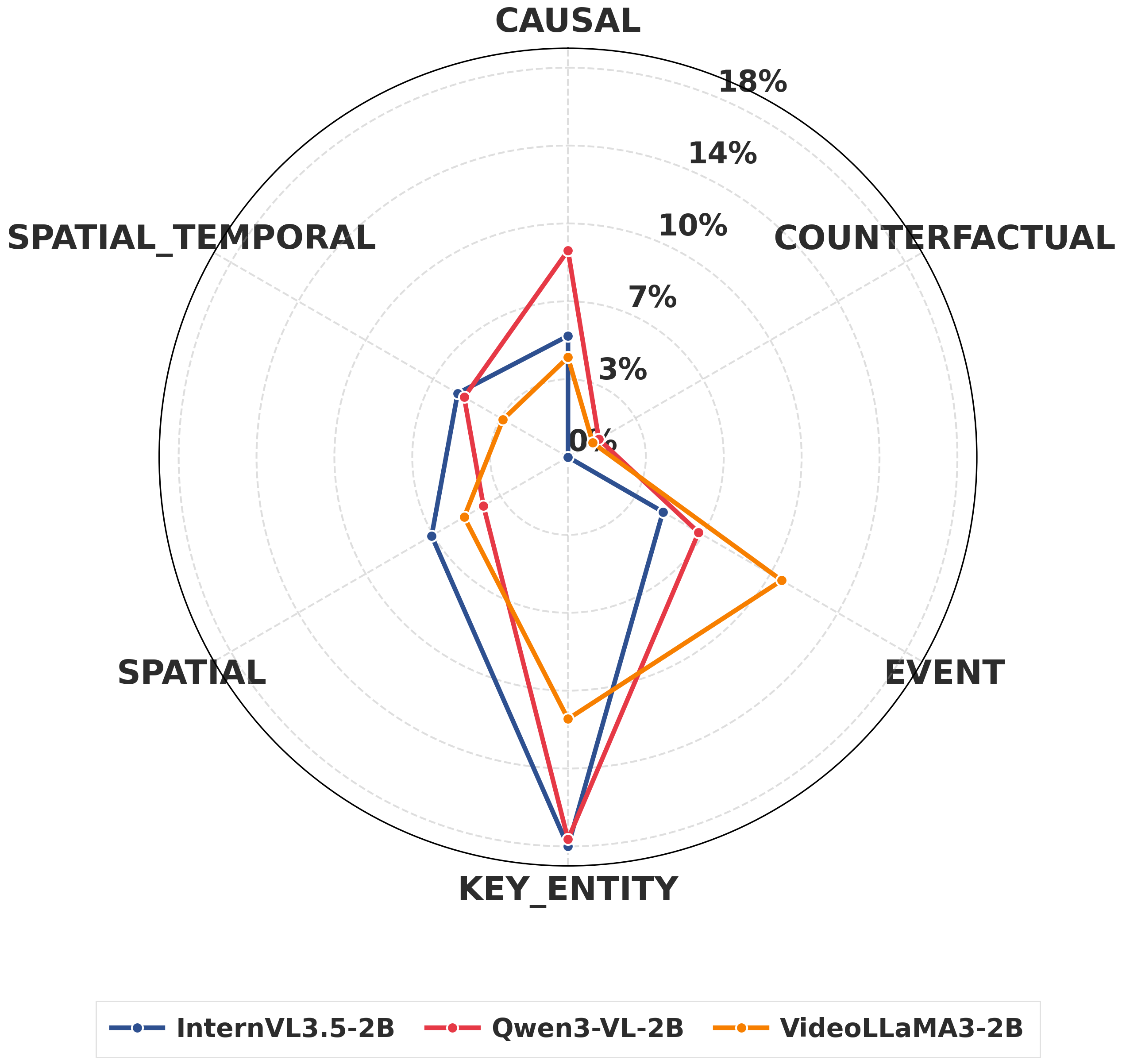}
    \caption{Small models}
  \end{subfigure}
  \hfill
  \begin{subfigure}[t]{0.31\textwidth}
    \centering
    \includegraphics[width=\textwidth]{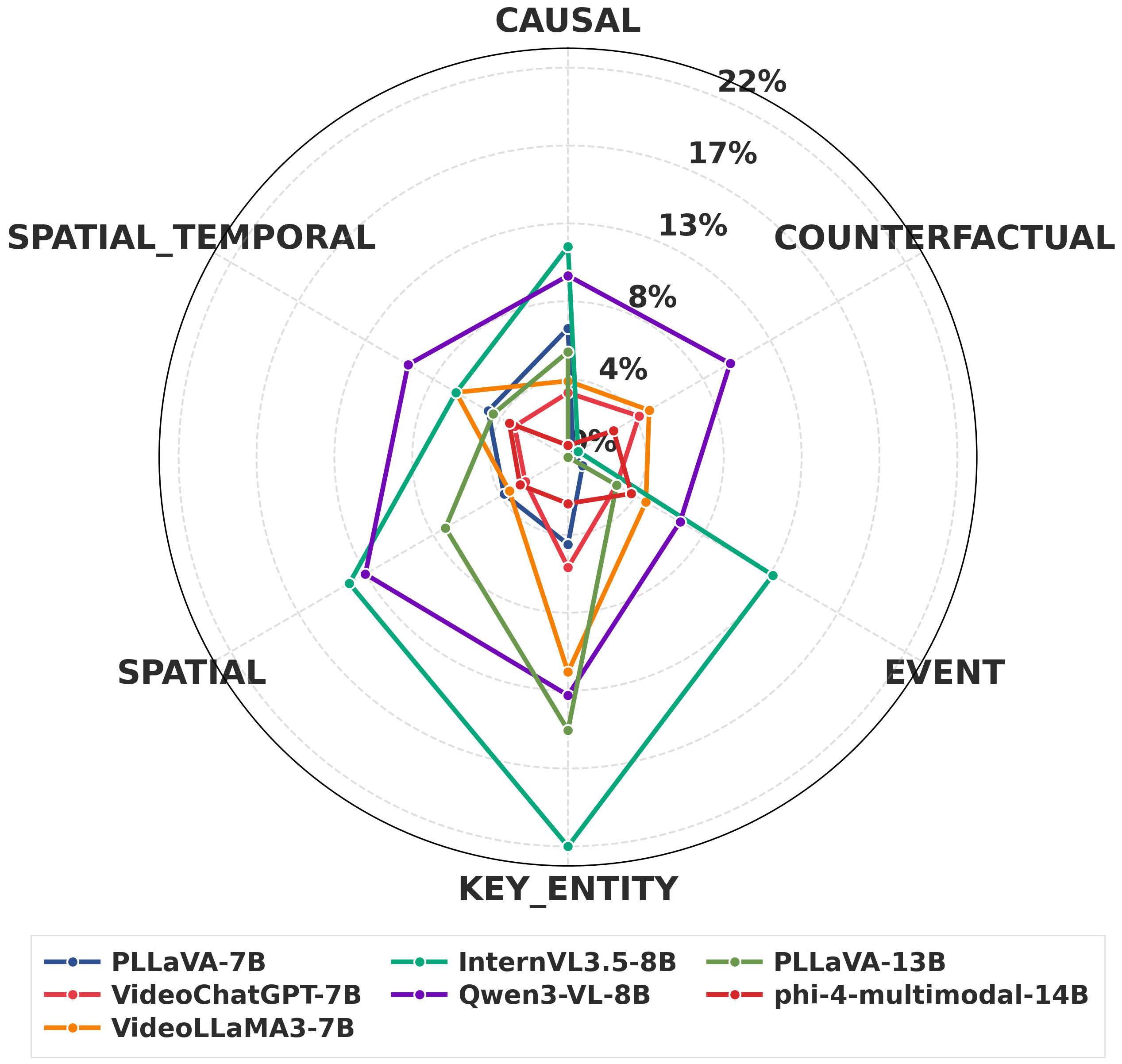}
    \caption{Medium models}
  \end{subfigure}
  \hfill
  \begin{subfigure}[t]{0.31\textwidth}
    \centering
    \includegraphics[width=\textwidth]{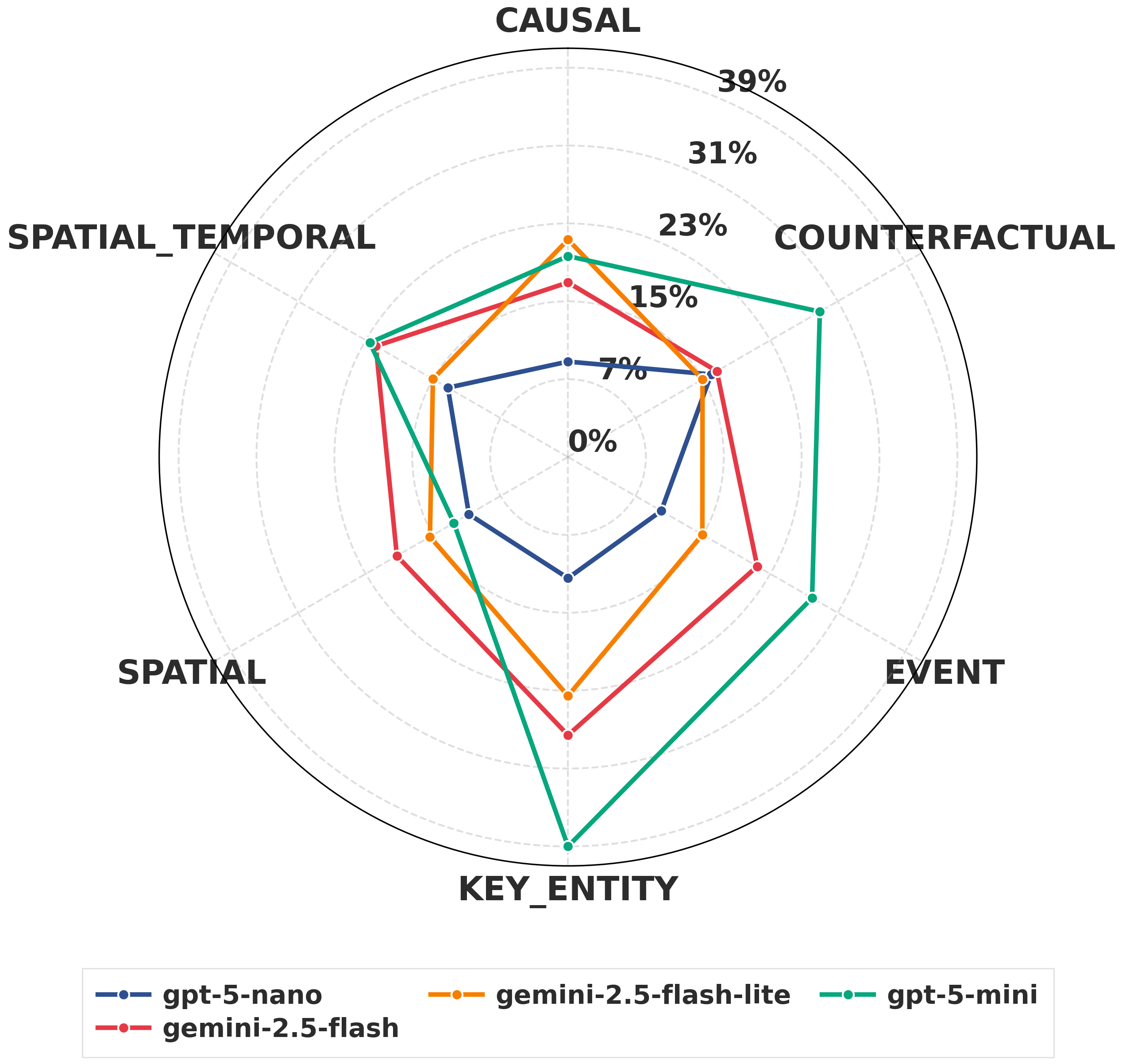}
    \caption{Large/commercial models}
  \end{subfigure}

  \caption{Category-wise QuadAcc radar plots for (a) small, (b) medium, and (c) large/commercial model groups.}
  \label{fig:quad_radar}
\end{figure*}

\subsection{Results and Analysis}

\paragraph{Statistical Reliability.}

To quantify statistical reliability under the benchmark scale, we perform scene-level bootstrap resampling with 2,000 replicates and report 95\% confidence intervals for all metrics. Across models, interval widths are typically around 1--2 percentage points, indicating low variance. For example, GPT-5-mini achieves a QuadAcc of 25.85\% with a 95\% CI of [24.32, 27.42], while its video consistency and question consistency are 58.10\% [57.07, 59.14] and 56.05\% [54.98, 57.12], respectively. We further evaluate ranking stability under bootstrap resampling and observe that model rankings remain consistent, with GPT-5-mini ranked first on QuadAcc across resampled datasets.

\paragraph{Large gap between global classification and contrastive consistency.}
Tab.~\ref{tab:quadacc_detailed} shows that most models reach moderate binary QA quality, with BaAcc commonly in the 50--70\% range and non-trivial MCCScore, yet QuadAcc remains low across the board.  For open-source models, QuadAcc is typically below 12\%, and even the best open-source entries only approach low double digits despite BaAcc around 60\% and above. The gap is also visible among proprietary models, which indicates that many models can produce plausible answers for isolated video question instances, but do not implement a stable decision rule that generalizes across the structured quadruple.

\begin{table}[t]
\centering
\small
\setlength{\tabcolsep}{4pt}
\caption{Vision-language sensitivity and binding diagnostics across models on \textsc{CCTVBench}. All values are percentages. Values are reported as mean percentages, with 95\% bootstrap confidence intervals estimated by bootstrap. Shaded cells indicate the best result within each model group, and boldface marks the best overall result in each column.}
\label{tab:metrics_main}
\resizebox{0.5\textwidth}{!}{%
\begin{tabular}{l c cccc}
\toprule
\multirow{2}{*}{\textbf{Model}} &
\multirow{2}{*}{\textbf{Size}} &
\multicolumn{4}{c}{\textbf{VL Sensitivity \& Binding Diagnostics}} \\
\cmidrule(lr){3-6}
 &  & VS$\uparrow$ & QS$\uparrow$ & GVRS$\uparrow$ & SVE$\uparrow$ \\
\midrule

\multicolumn{6}{c}{\textit{Open-source Models}} \\
\midrule
InternVL3.5      & 2B  & 16.05 & 29.92 & 20.79 & 10.69 \\
InternVL3.5      & 8B  & 27.75 & 36.11 & 31.27 & 14.41 \\
PLLaVA           & 7B  & 16.33 & 30.45 & 21.14 & 9.35 \\
PLLaVA           & 13B & 18.02 & 30.42 & 22.53 & 10.30 \\
Qwen3-VL         & 2B  & 20.34 & 29.08 & 23.82 & 11.01 \\
Qwen3-VL         & 8B  & 16.36 & 29.03 & 20.84 & 14.02 \\
VideoLLaMA3      & 2B  & \textbf{40.89} & 42.54 & 41.61 & \textbf{23.27} \\
VideoLLaMA3      & 7B  & \cellcolor{black!15}38.06 & \cellcolor{black!15}\textbf{48.13} & \cellcolor{black!15}\textbf{42.41} & \cellcolor{black!15}21.01 \\
VideoChatGPT     & 7B  & 31.92 & 31.61 & 31.67 & 16.63 \\
Phi-4-Multimodal & 14B & 19.08 & 34.36 & 24.42 & 13.41 \\
\midrule
\multicolumn{6}{c}{\textit{Proprietary Models}} \\
\midrule
Gemini-2.5 & flash-lite & 25.70 & \textbf{83.71} & 39.25 & 18.32 \\
Gemini-2.5 & flash      & 36.06 & 63.70 & 45.98 & \textbf{31.38} \\
GPT-5      & nano       & 33.00 & 40.68 & 36.36 & 28.68 \\
GPT-5      & mini       & \cellcolor{black!15} \textbf{36.67} & \cellcolor{black!15}68.67 & \cellcolor{black!15}\textbf{47.74} & \cellcolor{black!15}30.97 \\
\bottomrule
\end{tabular}%
}
\end{table}

\paragraph{Video and question consistency expose distinct weaknesses.}
The consistency breakdown in Tab.~\ref{tab:quadacc_detailed} reveals two common behaviors.
First, several models achieve very high Reject@q$^-$ but low Contr@q$^+$, meaning they reliably reject the mutually exclusive alternative yet frequently miss the true event on $v^+$.
Second, other models show stronger Contr@v$^+$ but weaker Reject@v$^-$, meaning they can separate q$^+$ from q$^-$ on the positive video but still produce excessive \emph{Yes} answers on the counterfactual video, violating the none-of-the-above requirement.

\paragraph{Failure-mode composition across models.}
Fig.~\ref{fig:failure_mode_stacked} decomposes failed quadruples into four failure patterns. 
Model size affects the omission-hallucination balance but not in a consistent direction across families.
Within GPT-5, the smaller variant becomes markedly more conservative and fails mainly by Positive Omission, while the larger variant shifts toward Negative Hallucination.
In Qwen3-VL, scaling to 8B also increases Positive Omission, suggesting that improved fluency or safety calibration can collapse decisions toward \texttt{No} rather than improving contrastive binding.  


\paragraph{High reactivity does not imply correct binding.}
Regarding sensitivity diagnostics, Tab.~\ref{tab:metrics_main} shows that strong modality sensitivity alone is insufficient.
VideoLLaMA3 attains the highest VS and QS, and also leads on SVE among open-source models, indicating that its predictions change substantially under controlled swaps.
Yet its QuadAcc in Tab.~\ref{tab:quadacc_detailed} remains limited, suggesting that the direction of these changes is often misaligned with the contrastive constraints.
Besides, GPT-5-mini combines high GVRS with the highest SVE and also achieves the best QuadAcc. Gemini-2.5-flash-lite shows extremely high QS but weaker QuadAcc, consistent with over-reacting to wording differences without reliably enforcing rejection on $v^-$.

\begin{table}[t]
\centering
\small
\setlength{\tabcolsep}{5pt}
\caption{Effect of contrastive decoding variants on classification and quadruple consistency. Subscripts denote signed change relative to Base.}
\label{tab:quadacc_grouped_final}
\resizebox{0.5\textwidth}{!}{%
\begin{tabular}{l l c c c}
\toprule
\textbf{Model} & \textbf{Decoding} &
\textbf{BaAcc}$\uparrow$ &
\textbf{Score}$_{\textbf{cls}}$$\uparrow$ &
\textbf{QuadAcc}$\uparrow$ \\
\midrule

\multirow{4}{*}{Qwen3-VL-2B}
& Base & 62.96 & 37.77 & 7.62 \\
& VCD  & $58.76_{\nega{-4.20}}$ & $33.19_{\nega{-4.58}}$ & $8.46_{\pos{+0.85}}$ \\
& TCD  & $63.56_{\pos{+0.60}}$ & $38.18_{\pos{+0.41}}$ & $7.56_{\nega{-0.06}}$ \\
& C-TCD
& \cellcolor{black!15}{$65.84_{\pos{+2.88}}$}
& \cellcolor{black!15}{$40.69_{\pos{+2.92}}$}
& \cellcolor{black!15}{$9.85_{\pos{+2.23}}$} \\


\midrule
\multirow{4}{*}{Qwen3-VL-8B}
& Base & 60.26 & 38.62 & 10.65 \\
& VCD  & $63.20_{\pos{+2.95}}$ & $39.69_{\pos{+1.06}}$ & $14.62_{\pos{+3.97}}$ \\
& TCD  & $61.82_{\pos{+1.56}}$ & $39.62_{\pos{+1.00}}$ & $12.70_{\pos{+2.05}}$ \\
& C-TCD
& \cellcolor{black!15}{$64.82_{\pos{+4.56}}$}
& \cellcolor{black!15}{$43.03_{\pos{+4.41}}$}
& \cellcolor{black!15}{$18.63_{\pos{+7.97}}$} \\



\bottomrule
\end{tabular}%
}
\end{table}

\paragraph{Category-wise analysis highlights persistent gaps in counterfactual and causal reasoning.}
Fig.~\ref{fig:quad_radar} summarizes QuadAcc by question category. Key-Entity tends to be the easiest, likely because it reduces to identifying salient participants and outcomes in accident scenes.
Event and Spatial categories are intermediate. Counterfactual and Causal remain consistently hard, because they require rejecting plausible hypotheses on counterfactual videos and resisting language priors that associate accident scenes with stereotypical causes. Proprietary models excel across all categories, but the radar shapes still show imbalance, with most models concentrating successes in a subset of categories rather than achieving uniform consistency.

\paragraph{Contrastive decoding benefits most from semantically decisive counterfactual pairs.}
Tab.~\ref{tab:quadacc_grouped_final} compares training-free contrastive decoding variants on Qwen3-VL.
VCD can improve QuadAcc but may reduce BaAcc and Score$_\text{cls}$, indicating that visually corrupted negatives can over-suppress useful evidence in strict Yes/No classification.
TCD yields smaller and less consistent gains. C-TCD improves all reported metrics simultaneously and produces the largest QuadAcc gains. These results suggest that using a semantically exclusive counterpart video provides a more targeted contrast signal than generic corruption or temporal degradation, aligning well with the benchmark's design.

\subsection{Discussion}
\label{sec:discussion}

\paragraph{Why QuadAcc is hard: enforcing none-of-the-above without collapsing to conservatism.}
The quadruple protocol is intentionally asymmetric: $q^+$ is true only on $v^+$, while both questions must be rejected on $v^-$.
Fig.~\ref{fig:failure_mode_stacked} shows that many models do not handle this none-of-the-above state robustly.
A common pattern is that rejecting $q^-$ is much easier than answering $q^+$ correctly. The other bottleneck is rejecting both questions on $v^-$. Even strong models with high Contr@v$^+$ still leak \emph{Yes} on the counterfactual, and lead to QuadAcc fail.

\paragraph{Language priors vs.\ visual grounding: imbalance shows up differently across metrics.}

Strong language grounding yields reasonable BaAcc, since many queries align with accident patterns and can be answered from text priors alone. However, when decisions are driven mainly by question wording and not by the video, QuadAcc remains low, because predictions fail to flip correctly across $v^+$ and $v^-$. In contrast, models with stronger visual grounding react more to video changes and can better separate positive and counterfactual scenes, which helps QuadAcc, but they still lose contrastive consistency if they do not distinguish mutually exclusive formulations. High BaAcc, therefore, does not imply high QuadAcc unless visual evidence and language understanding are jointly calibrated, as reflected by GVRS and SVE in Tab.~\ref{tab:metrics_main}.

\paragraph{Implications: Counterfactual world modeling is useful beyond evaluation.}
The consistent gains of C-TCD indicate that counterfactual pairs can serve as a strong inference-time anchor. More broadly, they suggest an interesting direction: align models using paired \((v^+, v^-)\) with objectives that explicitly enforce mutual exclusivity on \(v^+\), rather than relying solely on single-video QA supervision. It also shows that driving world models facilitates model consistency ability, leveraging contrastive video pairs.

\paragraph{Benchmark scope.}
CCTVBench is a controlled contrastive benchmark rather than a large-scale distributional dataset. It focuses on event-level paired scenes with minimal counterfactual edits, preserving causal structure but limiting coverage of long-tail driving scenarios. Compared to existing traffic VideoQA benchmarks that emphasize large-scale QA under natural distributions, CCTVBench provides a complementary setting for evaluating consistency under near-identical counterfactual conditions, trading coverage for diagnostic precision.

\section{Conclusion}
\label{sec:conclusion}

We introduced \textsc{CCTVBench}, a contrastive traffic VideoQA benchmark built from real accident footage paired with generated counterfactual normal videos. The contrastive quadruple protocol enforces a single coherent decision rule and explicitly probes calibrated rejection under minimal scene changes for risk-aware reasoning. We also provide actionable diagnostics that separate cross-video and cross-question consistency and attribute failures to 4 failure patterns. Experiments across diverse video LLMs reveal a persistent gap between per-instance classification scores and quadruple-level contrastive consistency. We further proposed C-TCD, showing that semantically contrastive video pairs serve not only as an evaluation construct but also as an effective inference-time contrast signal that improves consistency without retraining.

\section*{Limitations}

CCTVBench has several limitations. First, counterfactual negatives are generated using a single driving world model, as our setting requires temporally consistent video-conditioned generation with stable camera geometry, which is not supported by most publicly accessible alternatives. Although we mitigate artifacts through filtering and human review, some generator-specific bias may remain. Second, the benchmark scale is constrained by the cost of collecting rare accident scenes and constructing paired counterfactuals, limiting statistical coverage and external validity. Third, ensuring strictly mutually exclusive yet not collectively exhaustive video-text pairs is challenging, and some residual ambiguity may persist despite manual validation.

\clearpage
\bibliography{custom}

\appendix

\clearpage

\section{Appendix}


\begin{table*}[h!]
\centering
\caption{\textbf{Comparison of faithfulness-oriented and traffic VQA benchmarks.}
\textnormal{
\#Ques = \#QA pairs ; \#Img/\#Video = unique images / video clips.
Eval: LLM = LLM-judge; B-QA = binary QA; MCQ = multiple-choice QA; OE = open-ended.
CP (contrastive pairing): Video CP = contrastive/paired videos; Text CP = contrastive/paired questions.
}}
\resizebox{\linewidth}{!}{
\begin{tabular}{lcccccccc}
\toprule
\textbf{Benchmark} & \textbf{Modality} &
\textbf{\#Ques} & \textbf{\#Img} & \textbf{\#Video} &
\textbf{Eval} &
\textbf{Video CP} & \textbf{Text CP} \\
\midrule
\multicolumn{8}{c}{\textit{Faithfulness VQA benchmarks}} \\
\midrule
MMHal-Bench \cite{sun2023llavarlhf} & Image & 96 & 96 & - & LLM & \xmark & \xmark \\
EasyDetect \cite{chen-etal-2024-unified-hallucination} & Image & 420 & 420 & - & LLM & \xmark & \xmark \\

VideoHallucer \cite{wang2024videohallucer} & Video & 1,800 & - & 948 & B-QA/OE & \xmark & \cmark \\
EventHallusion \cite{zhang2025eventhallusiondiagnosingeventhallucinations} & Video & 711 & - & 400 & B-QA/OE & \xmark & \xmark \\
VALOR \cite{qiu-etal-2024-valor} & Image & 211 & 211 & - & LLM & \xmark & \cmark \\

HallusionBench \cite{guan2023hallusionbench} & Image, Video & 1,129 & - & 346 & LLM & \xmark & \cmark \\
CertainlyUncertain \cite{chandu2024certainlyuncertainbenchmarkmetric} & Image & 178,000 & - & 95,800 & VQA & \xmark & \cmark \\

EGOILLUSION \cite{seth2025egoillusionbenchmarkinghallucinationsegocentric} & Video & 8,000 & - & 1,400 & Closed+Open & \xmark & \xmark \\
VidHalluc \cite{li2025vidhalluc} & Video & 9,295 & - & 5,002 & B-QA/MCQ/OE/Sort & \cmark & \cmark \\
\midrule
\multicolumn{8}{c}{\textit{Traffic VQA benchmarks}} \\
\midrule
SUTD-TrafficQA \cite{xu2021sutdtrafficqa} & Video & 62,535 & - & 10,080 & MQA & \xmark & \xmark \\
TUMTraffic-VideoQA \cite{zhou2025tumtrafficvideoqa} & Video & 85,000 & - & 1,000 & MCQ & \xmark & \xmark \\
NuScenes-QA \cite{qian2024nuscenesqa} & Multi-view Image, LiDAR & 460,000 & 34,000 & - & VQA & \xmark & \xmark \\
DriveBench \cite{DrivingBenchxie2025vlmsreadyautonomousdriving} & Image & 20,498 & 19,200 & - & VQA & \xmark & \xmark \\
STSBench \cite{fruhwirthreisinger2025stsbenchspatiotemporalscenariobenchmark} & Video & 971 & - & 43 & VQA & \xmark & \xmark \\
\midrule
\textbf{CCTVBench (Ours)} & Video & 7,104 & - & 610 & B-QA & \cmark & \cmark \\
\bottomrule
\end{tabular}}
\label{tab:faithfulness_traffic_bench}
\end{table*}


\subsection{Additional Related Work}

Recent work has moved beyond generic VQA accuracy to explicitly probe faithfulness, meaning answers must be grounded in visual evidence, and to diagnose hallucination. Early image-centric benchmarks ~\cite{sun2023llavarlhf,chen-etal-2024-unified-hallucination,qiu-etal-2024-valor} typically evaluate models via LLM-based judging or curated prompts, focusing on object, attribute, and relation errors. A clear trend is the use of paired or controlled questions for consistency evaluation under minimal linguistic perturbations \cite{guan2023hallusionbench, chandu2024certainlyuncertainbenchmarkmetric}, which contrastive QA settings that expose strong language priors and poor calibration on unanswerable cases. Video-based domain is less explored but rapidly emerging ~\cite{wang2024videohallucer,zhang2025eventhallusiondiagnosingeventhallucinations,seth2025egoillusionbenchmarkinghallucinationsegocentric}. However, most faithfulness benchmarks remain single-image or single-video centered. They measure whether a model answers each query correctly, but do not enforce a global decision pattern across mutually exclusive conditions.

As summarized in Tab.~\ref{tab:faithfulness_traffic_bench}, most existing video faithfulness datasets still lack scene-matched counterfactual video pairs that differ only in the critical event, which is essential to disentangle genuine visual grounding from plausible narrative completion. In parallel, traffic and autonomous driving VQA benchmarks have grown in scale and diversity ~\cite{xu2021sutdtrafficqa,zhou2025tumtrafficvideoqa,qian2024nuscenesqa,DrivingBenchxie2025vlmsreadyautonomousdriving,fruhwirthreisinger2025stsbenchspatiotemporalscenariobenchmark}. These benchmarks primarily emphasize capability in scene understanding, multi-agent reasoning, and planning-related queries via multiple-choice or mixed-format VQA, but they are not designed for contrastive faithfulness evaluation.


\subsection{Data Distribution and Statistics}
\label{app:dataset_stats}

\textsc{CCTVBench} contains 7,104 binary video-question instances, comprising 305 positive videos and 305 paired generated negative videos. The benchmark is organized into 1,776 complete contrastive quadruples, where each quadruple includes four instances $(v^+, q^+)$, $(v^+, q^-)$, $(v^-, q^+)$, and $(v^-, q^-)$. All samples fall into six categories: \texttt{event}, \texttt{key\_entity}, \texttt{spatial}, \texttt{spatial\_temporal}, \texttt{causal}, and \texttt{counterfactual}. The dataset is strictly paired with equal numbers of positive and negative videos and has 100\% quadruple coverage.

\begin{figure}[ht]
  \centering
  \begin{subfigure}[t]{0.48\textwidth}
    \centering
    \includegraphics[width=\textwidth]{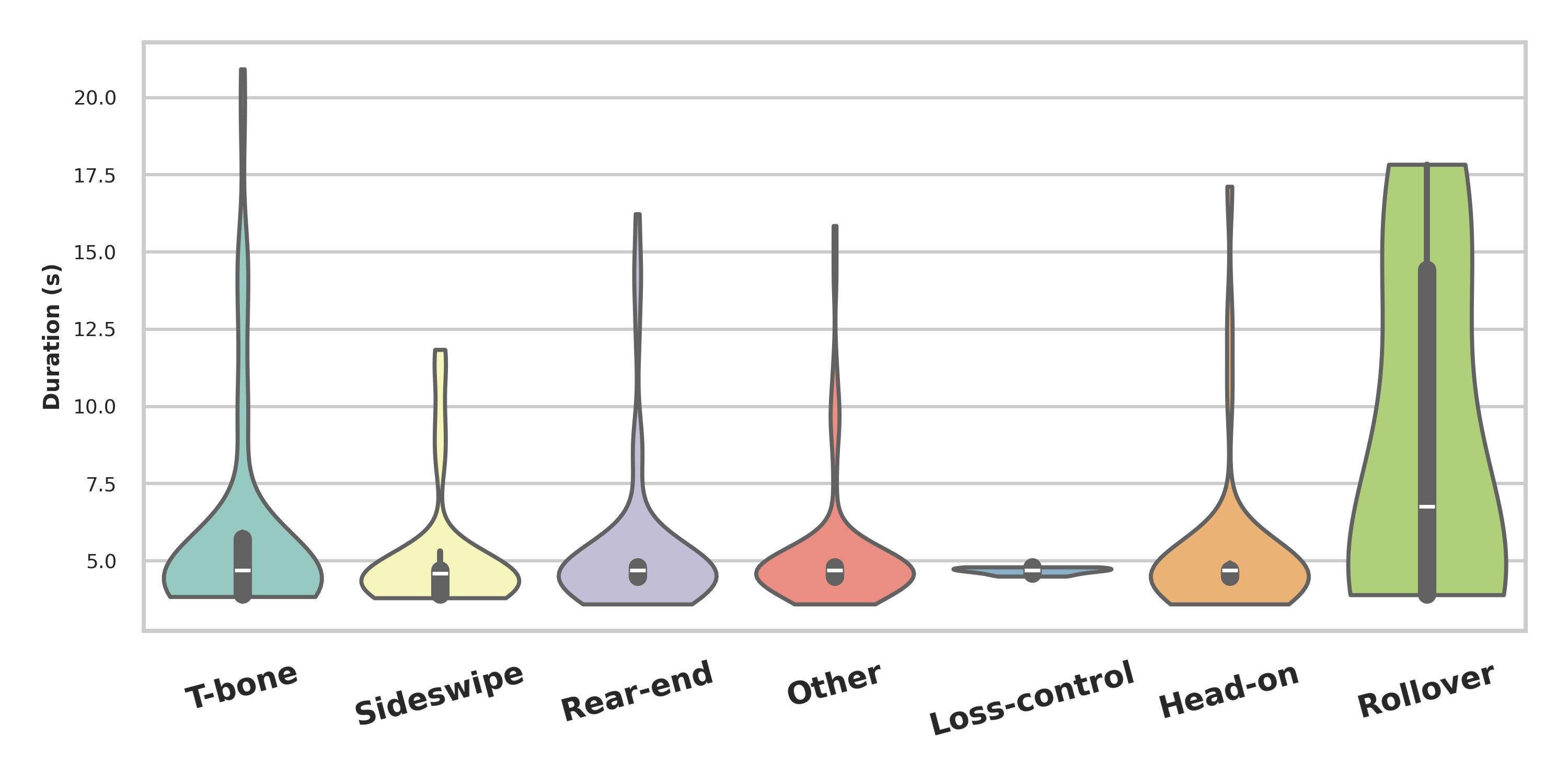}
    \caption{ Clip duration distributions across major types}
  \end{subfigure}
  \begin{subfigure}[t]{0.46\textwidth}
    \centering
    \includegraphics[width=\textwidth]{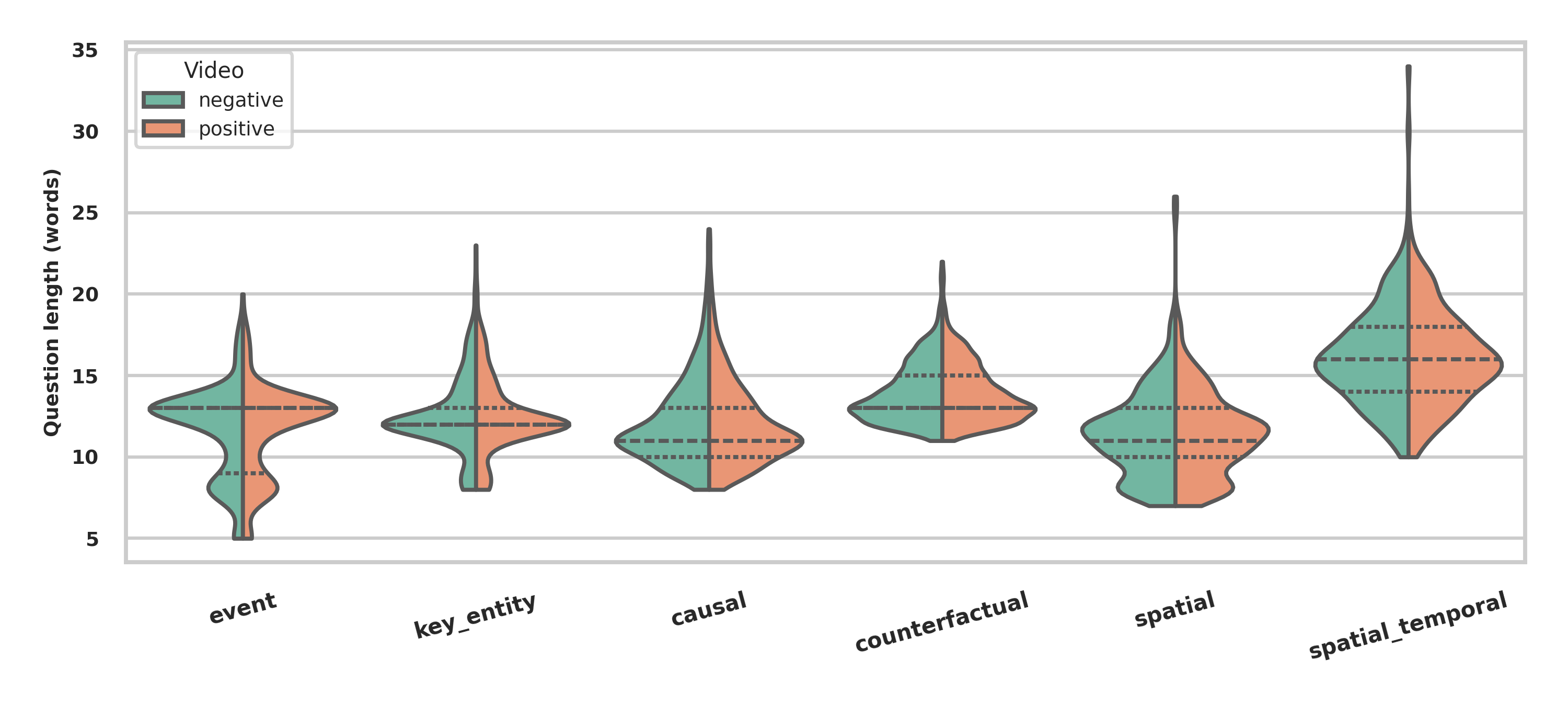}
    \caption{Question length distributions by category}
  \end{subfigure}
  \begin{subfigure}[t]{0.25\textwidth}
    \centering
    \includegraphics[width=\textwidth]{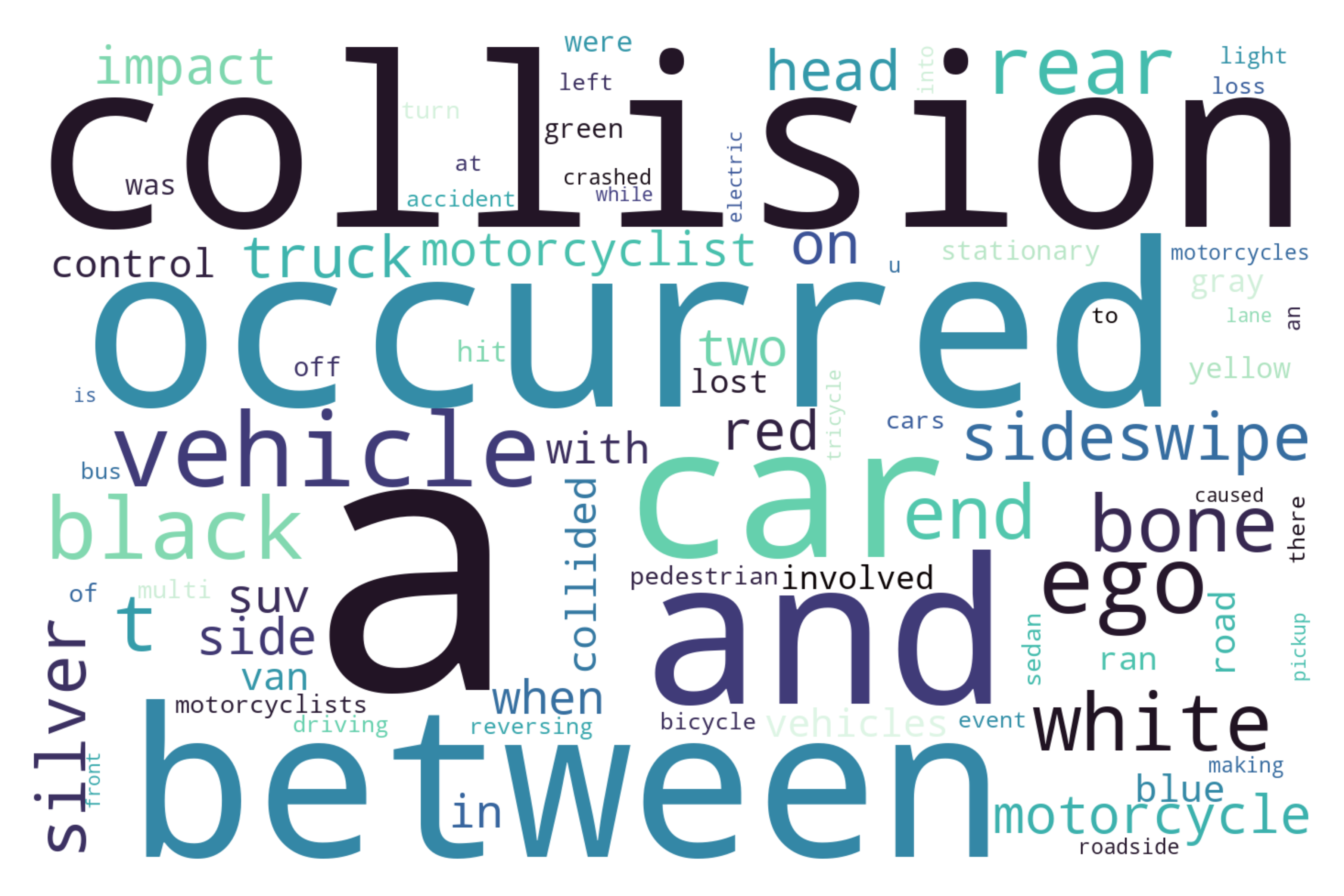}
    \caption{Word cloud distribution of traffic event caption}
  \end{subfigure}
  \hfill
\caption{CCTVBench dataset statistics. }
\label{fig:dataset_distribution}
\end{figure}

Fig.~\ref{fig:dataset_distribution} summarizes several basic properties of \textsc{CCTVBench}. 
Fig.~\ref{fig:dataset_distribution} (a) plot reports the distribution of clip durations across major anomaly types, showing that most clips concentrate around a short time window with a long tail of longer scenarios. 
 Fig.~\ref{fig:dataset_distribution} (b) plot compares question length across the six categories, indicating that the paired questions are concise and largely consistent in length, with moderate variation depending on whether the category emphasizes entities, spatial relations, or temporal descriptions. 
 Fig.~\ref{fig:dataset_distribution} (c) plot visualizes the overall question vocabulary as a word cloud, which is dominated by accident-centric terms such as \textit{collision}, and \textit{vehicle}.

\subsection{Dataset Generation Details}
\label{app:dataset-generation}

\textsc{CCTVBench} is constructed via a hybrid pipeline that couples (i) structured traffic video annotations, (ii) world-model-based counterfactual video synthesis to obtain video pairs, and (iii) LLM-assisted contrastive caption and QA generation.

\paragraph{Manual meta-data annotation.}
We annotate each positive clip $v_s^+$ with visual instance tracks and structured scene meta data.
We first run RT-DETR-L \cite{lv2024rtdetrv2improvedbaselinebagoffreebies} at 1~fps and ByteTrack \cite{zhang2022bytetrack} to obtain coarse trajectories, which annotators refine into consistent instance identities and key participants, along with hypothesized causes and critical spatio-temporal relations. Annotators also segment $v_s^+$ into base and contrastive segment. The resulting track IDs, event segments, and motion cues constrain downstream caption and QA synthesis during LLM-assisted QA generation.

\paragraph{Counterfactual video synthesis.}

To construct minimally different yet semantically decisive video pairs, we leverage NVIDIA Cosmos Predict-2 Video2World-2B to synthesize a counterfactual video $v_s^-$ in which the accident does not occur while preserving layout, illumination, and camera viewpoint as much as possible.
We condition generation on the preserved prefix using a small set of conditioning frames, typically five, and falling back to one when the prefix is short. We execute generation at 720p by default, and use 16 frames per second for videos with sufficiently high frame rates and 10 frames per second otherwise. We show view-specific safe prompts in Fig.~\ref {lst:safe_prompts} and negative prompt in Fig.~\ref {lst:negative_prompt}.

\begin{figure}[ht]
\centering
\begin{tcolorbox}[
  colframe=gray!75!black,
  colback=gray!3,
  boxrule=0.6pt,
  left=1mm,right=1mm,top=0.5mm,bottom=0.5mm,
  title={Safe prompts}
]
\begin{lstlisting}[
  basicstyle=\ttfamily\footnotesize,
  breaklines=true,
  breakatwhitespace=true,
  columns=fullflexible,
  keepspaces=true,
  showstringspaces=false
]
SAFE_PROMPTS = {
    "driving": [
        "Continue the driving scene with smooth, normal traffic flow. "
        "Show vehicles moving safely and predictably without any accidents, "
        "collisions, or traffic violations.",
    ],
    "roadside": [
        "Continue the roadside view showing normal traffic passing by safely. "
        "Maintain the fixed camera position with smooth vehicle movements and "
        "no traffic incidents. The objects in the scene remain in the same "
        "relative positions",
    ]
}
\end{lstlisting}
\end{tcolorbox}
\caption{Safe prompts for counterfactual video synthesis}
\label{lst:safe_prompts}
\end{figure}

\begin{figure}[ht]
\centering
\begin{tcolorbox}[
  colframe=gray!75!black,
  colback=gray!3,
  boxrule=0.6pt,
  left=1mm,right=1mm,top=0.5mm,bottom=0.5mm,
  title={Negative prompt}
]
\begin{lstlisting}[
  basicstyle=\ttfamily\footnotesize,
  breaklines=true,
  breakatwhitespace=true,
  columns=fullflexible,
  keepspaces=true,
  showstringspaces=false
]
NEGATIVE_PROMPT = (
    "reckless behavior, traffic jam, emergency vehicles, police chase, road rage, speeding, illegal maneuvers, "
    "traffic anomaly, traffic anomalies, anomaly, anomalies, inconsistency, inconsistencies, temporal inconsistency, temporal artifacts"
)
\end{lstlisting}
\end{tcolorbox}
\caption{Negative prompt for counterfactual video synthesis.}
\label{lst:negative_prompt}
\end{figure}

For long videos, we perform auto-regressive multi-segment generation with a maximum segment length of 5 seconds and re-condition each segment on the last conditioning frames from the previous segment. Prompts are selected based on the camera view, driving or roadside, to encourage normal continuation without collisions or violations, while a negative prompt list suppresses failure modes such as reckless behavior, anomalies, and temporal artifacts. After synthesis, we remove duplicated conditioning frames at segment boundaries, normalize the two videos to a consistent frame rate and resolution, and match their durations by truncating both videos to the shorter one, yielding temporally aligned pairs for contrastive evaluation.

\subsubsection{Category-wise Breakdown of Main Results}
\label{sec:appendix_category_breakdown}

\begin{figure*}[h]
  \centering
  \begin{subfigure}[t]{0.3\textwidth}
    \centering
    \includegraphics[width=\textwidth]{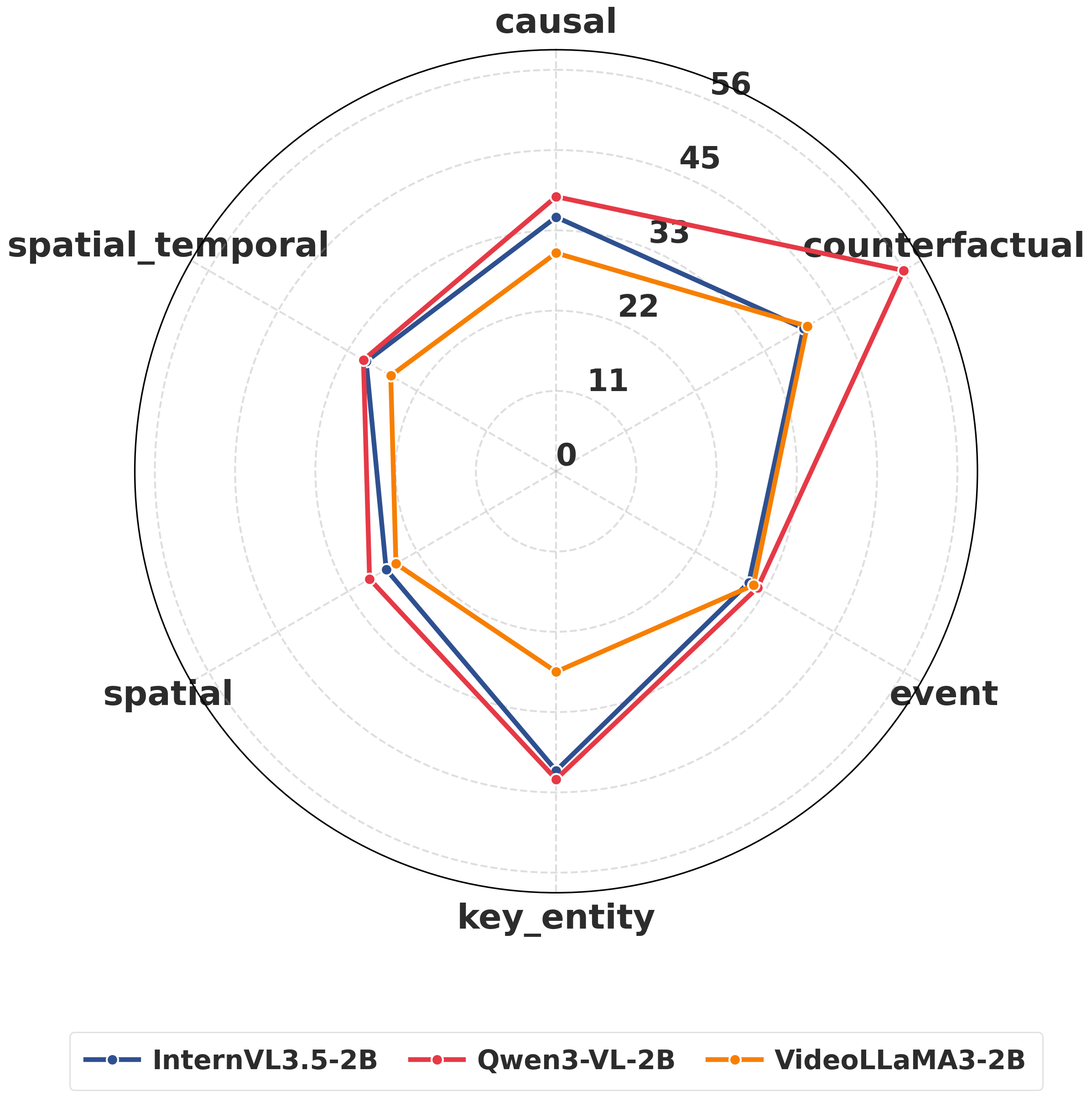}
    \caption{Small models}
    \label{fig:scorecls_radar_small}
  \end{subfigure}
  \hfill
  \begin{subfigure}[t]{0.3\textwidth}
    \centering
    \includegraphics[width=\textwidth]{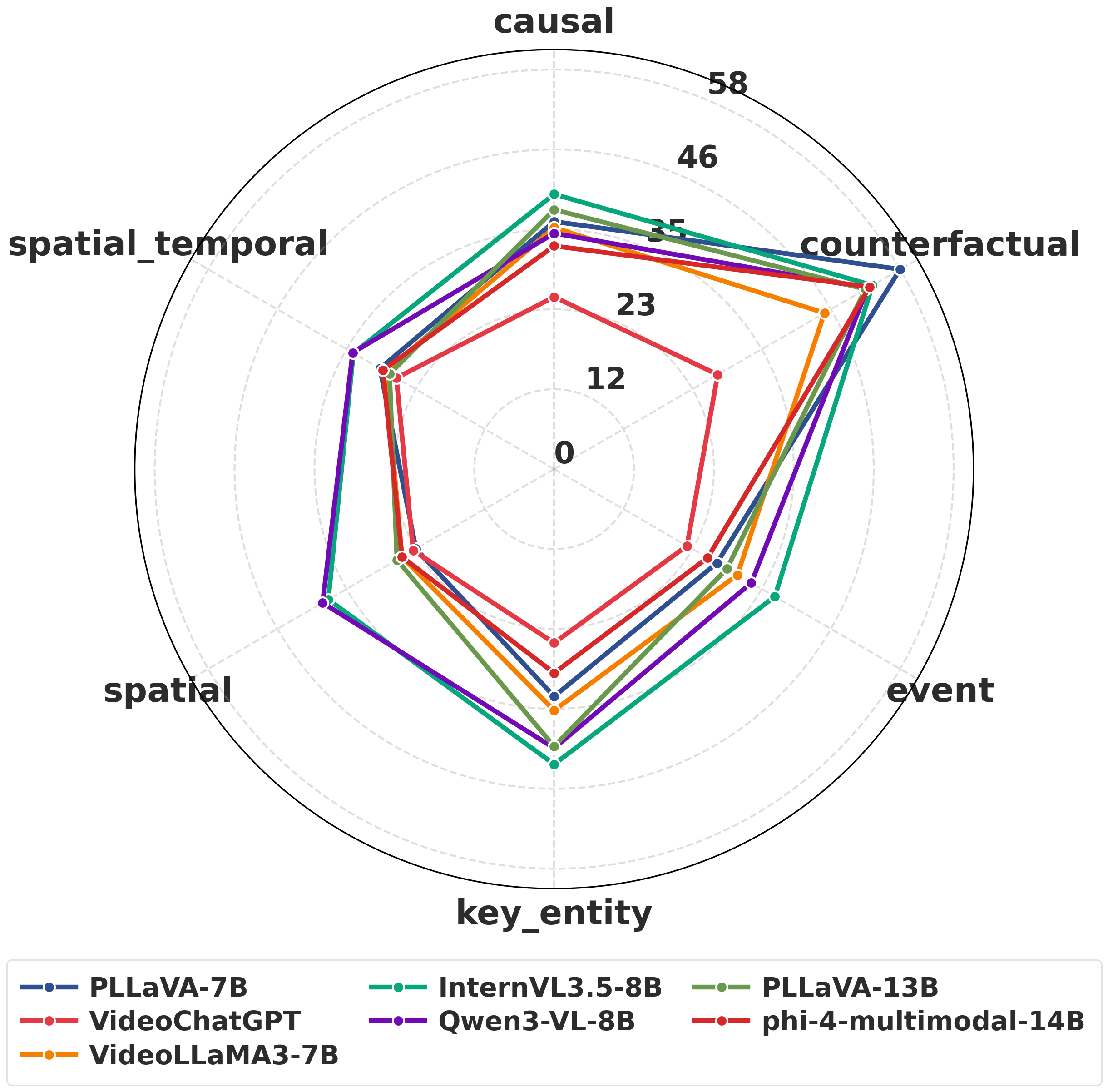}
    \caption{Medium models}
    \label{fig:scorecls_radar_medium}
  \end{subfigure}
  \hfill
  \begin{subfigure}[t]{0.3\textwidth}
    \centering
    \includegraphics[width=\textwidth]{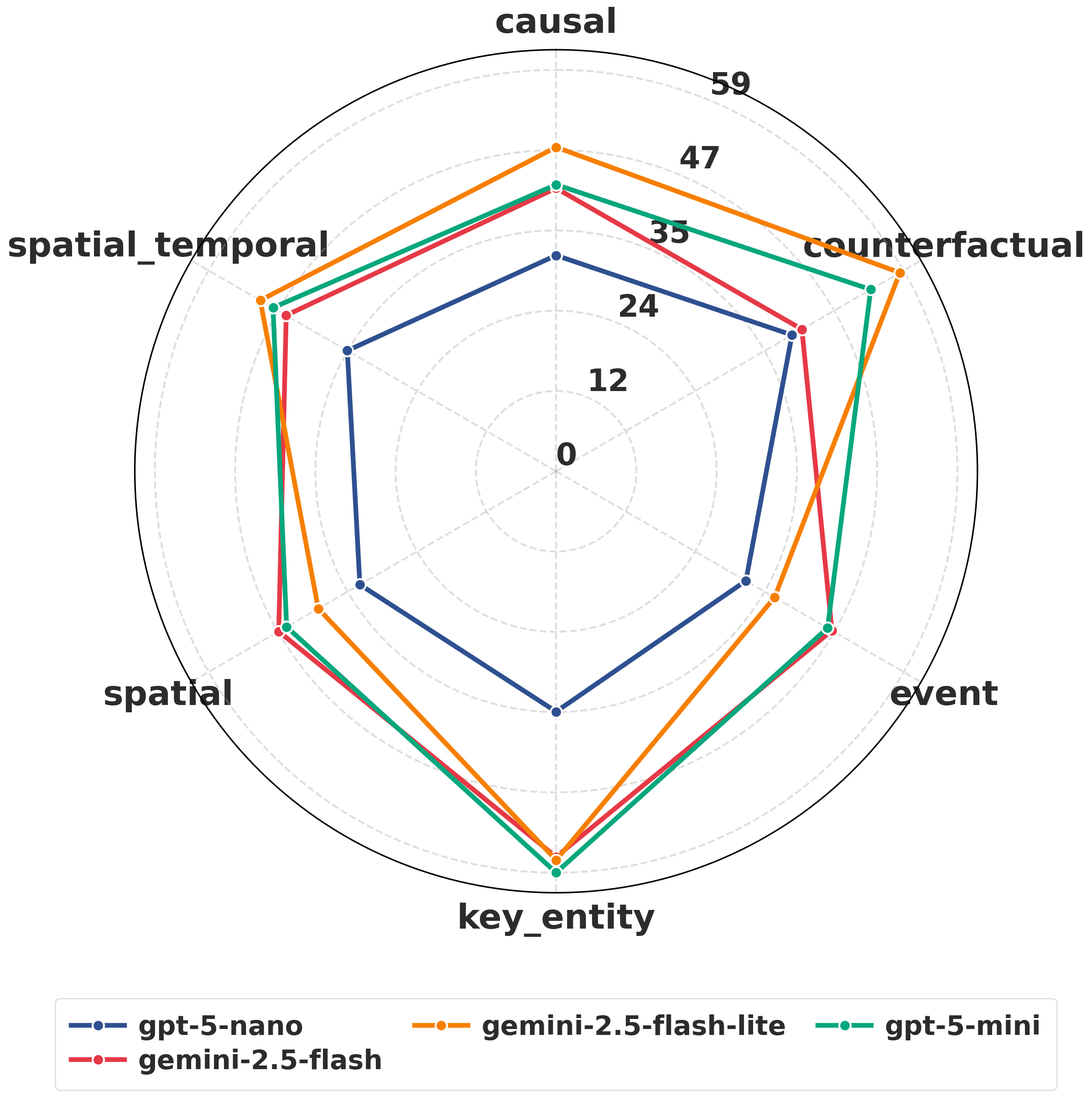}
    \caption{Large/commercial models}
    \label{fig:scorecls_radar_large}
  \end{subfigure}
  \caption{Category-wise classification score (Score$_{\mathrm{cls}}$) radar plots for small, medium, and large/commercial model groups.}
  \label{fig:scorecls_radar}
\end{figure*}

\begin{figure*}[h]
  \centering
  \begin{subfigure}[t]{0.3\textwidth}
    \centering
    \includegraphics[width=\textwidth]{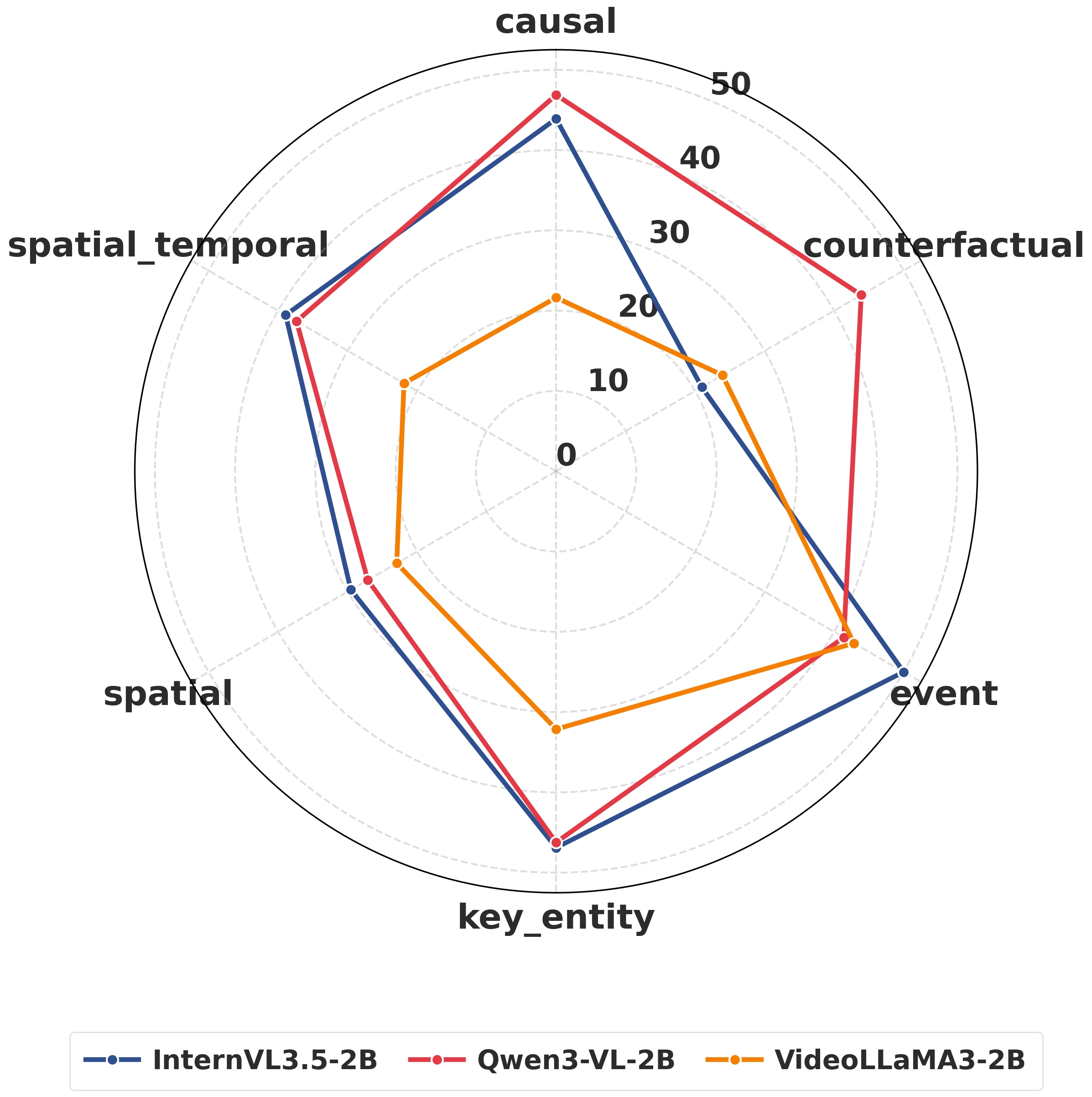}
    \caption{Small models}
    \label{fig:qcons_radar_small}
  \end{subfigure}
  \hfill
  \begin{subfigure}[t]{0.3\textwidth}
    \centering
    \includegraphics[width=\textwidth]{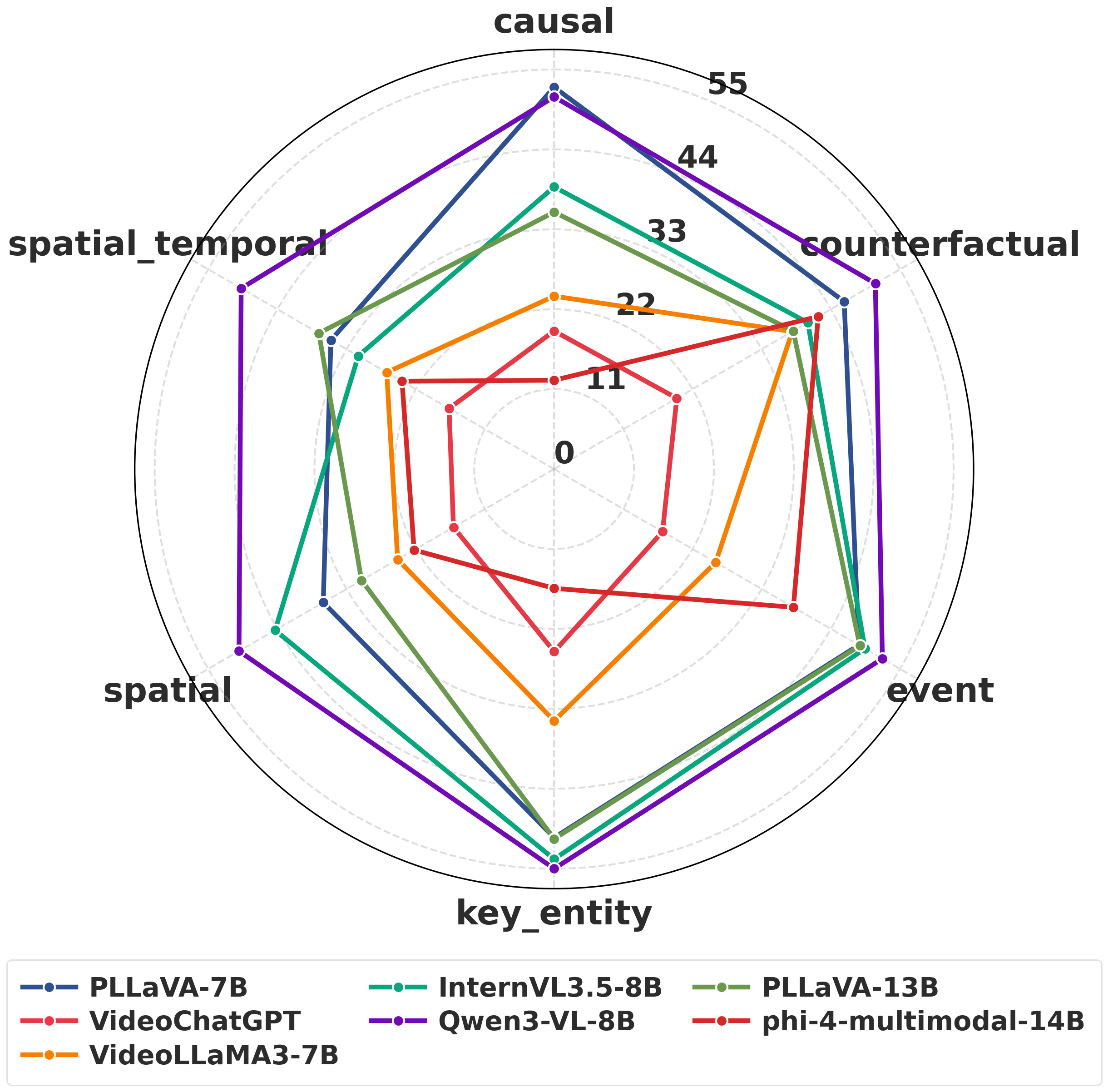}
    \caption{Medium models}
    \label{fig:qcons_radar_medium}
  \end{subfigure}
  \hfill
  \begin{subfigure}[t]{0.3\textwidth}
    \centering
    \includegraphics[width=\textwidth]{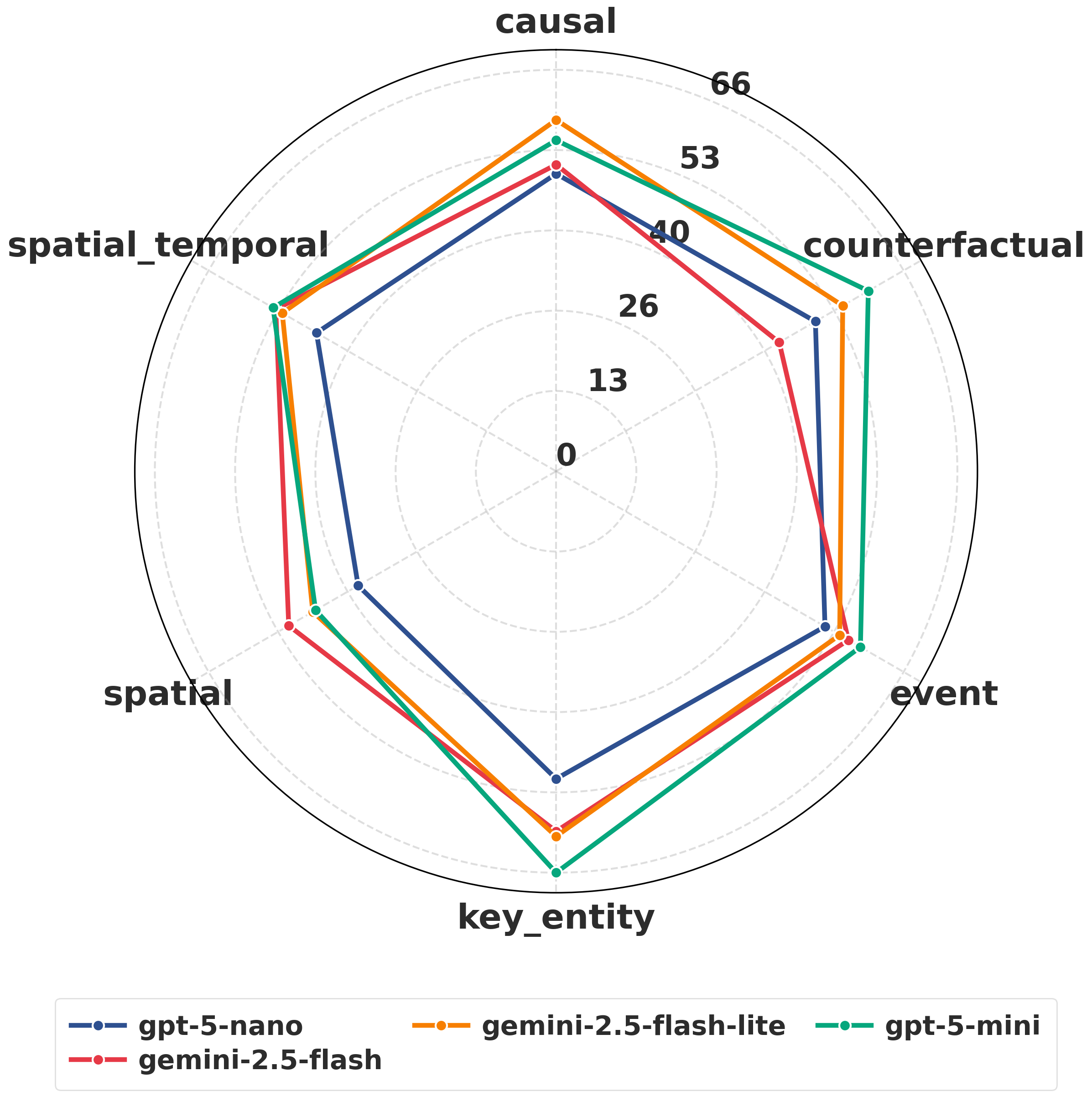}
    \caption{Large/commercial models}
    \label{fig:qcons_radar_large}
  \end{subfigure}
   \caption{Category-wise question consistency radar plots for small, medium, and large/commercial model groups.}
  \label{fig:qcons_radar}
\end{figure*}

\begin{figure*}[h!]
  \centering
  \begin{subfigure}[t]{0.3\textwidth}
    \centering
    \includegraphics[width=\textwidth]{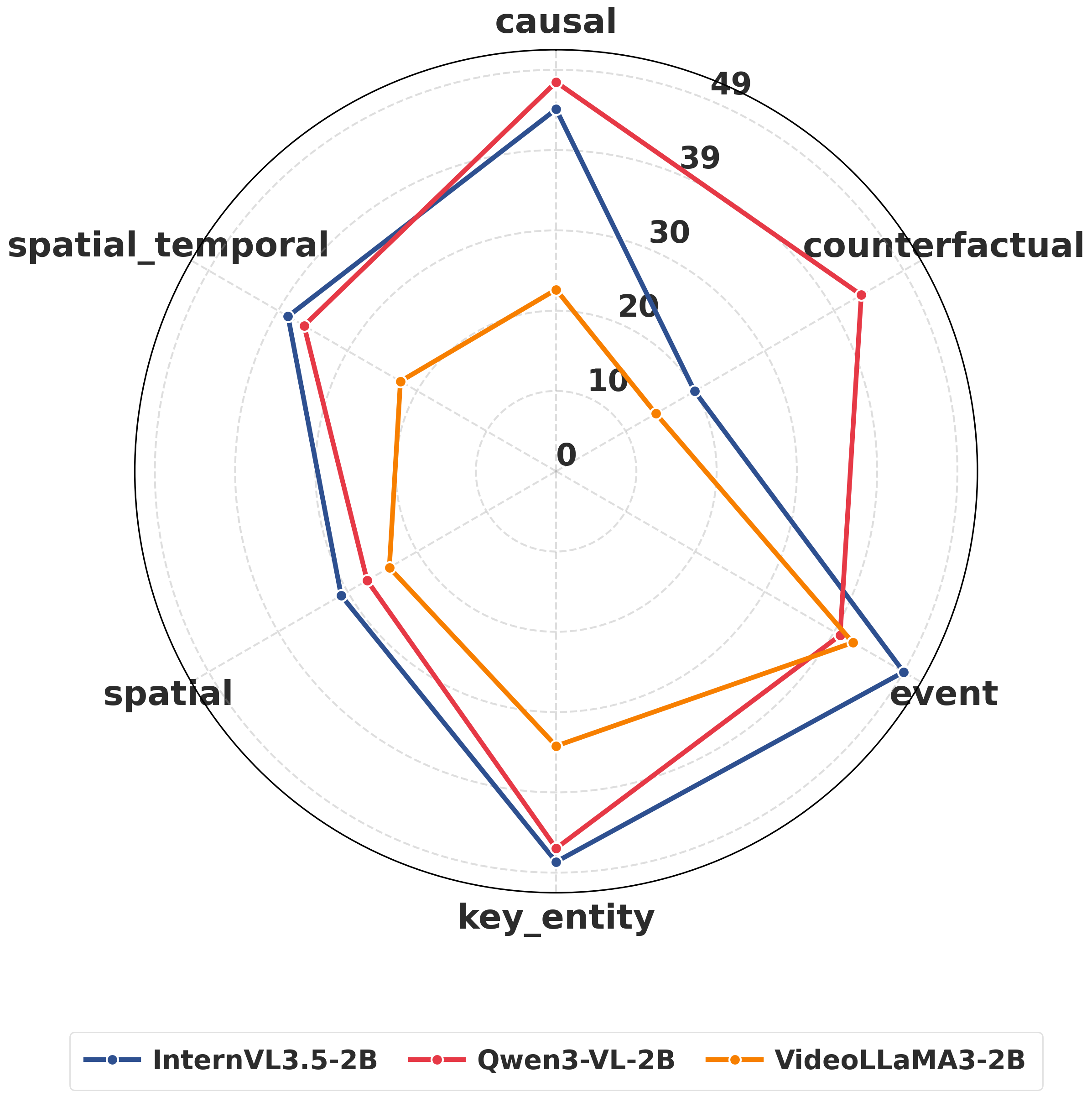}
    \caption{Small models}
    \label{fig:vcons_radar_small}
  \end{subfigure}
  \hfill
  \begin{subfigure}[t]{0.3\textwidth}
    \centering
    \includegraphics[width=\textwidth]{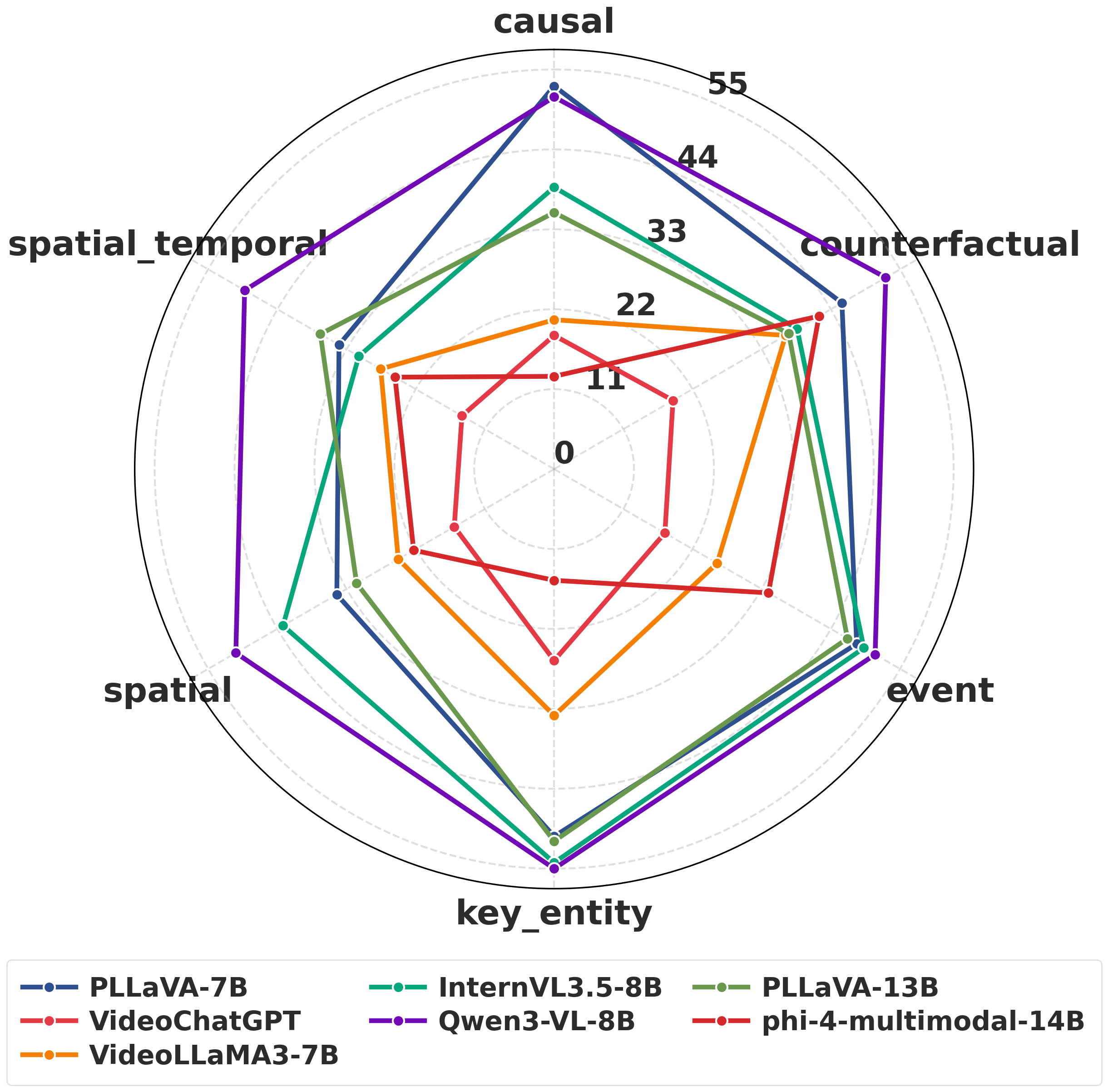}
    \caption{Medium models}
    \label{fig:vcons_radar_medium}
  \end{subfigure}
  \hfill
  \begin{subfigure}[t]{0.3\textwidth}
    \centering
    \includegraphics[width=\textwidth]{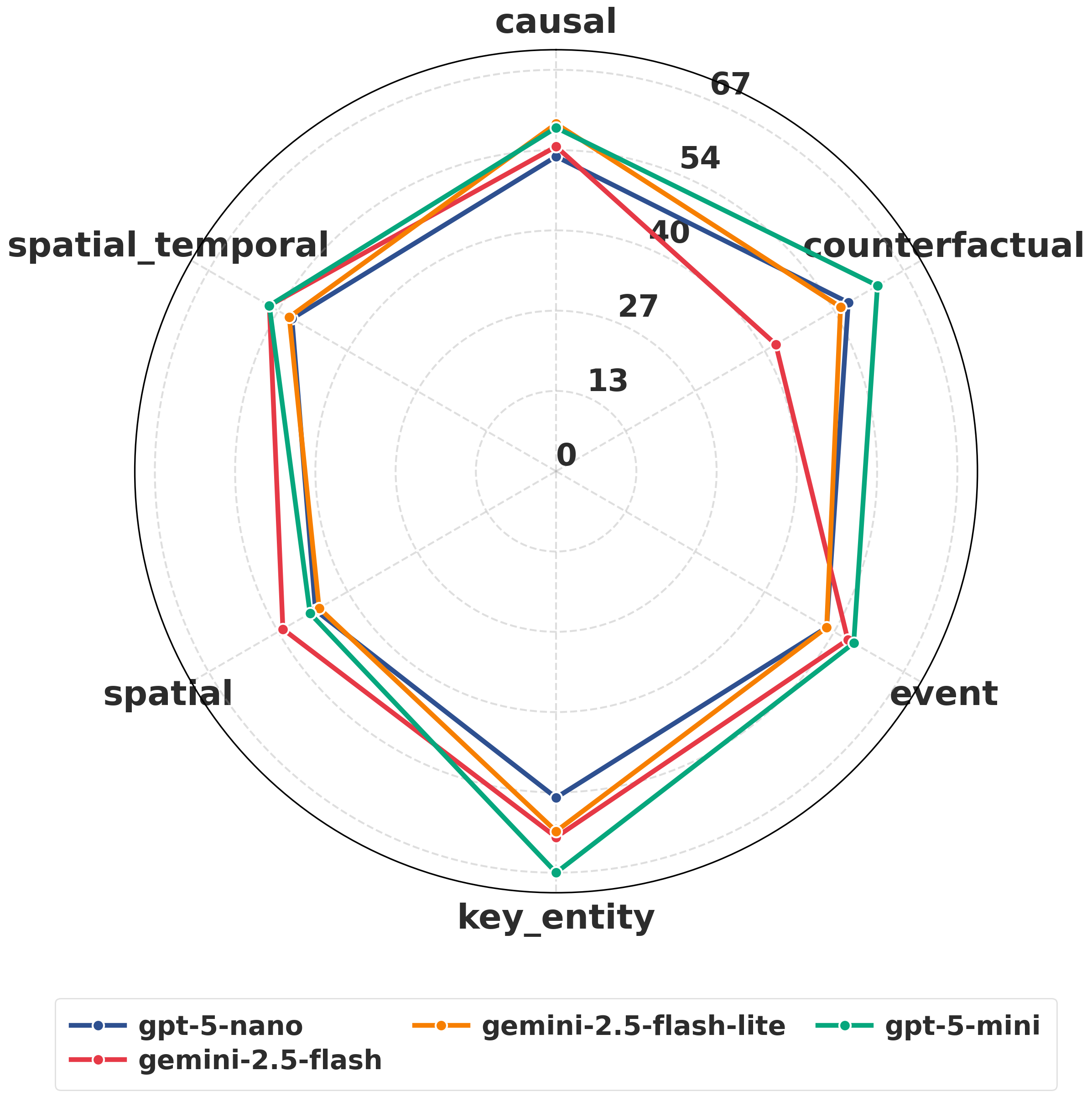}
    \caption{Large/commercial models}
    \label{fig:vcons_radar_large}
  \end{subfigure}
  \caption{Category-wise video consistency radar plots for small, medium, and large/commercial model groups.}
  \label{fig:vcons_radar}
\end{figure*}

Fig.~\ref{fig:scorecls_radar} summarizes the category-wise classification score Score$_{\mathrm{cls}}$. Across model groups, BaAcc is comparatively high and shows a relatively smooth profile across Event, Spatial, Spatial-Temporal, and Key-Entity, indicating that many models can reach moderate instance-level QA quality when evaluated independently. However, the consistency-oriented views in Fig.~\ref{fig:qcons_radar} and Fig.~\ref{fig:vcons_radar} expose much sharper category gaps and model separations. Key-Entity is the most stable category across groups, while Causal and Counterfactual remain the most challenging and also exhibit the largest variance across models, consistent with the benchmark requirement to reject plausible-but-false hypotheses under near-identical counterfactual scenes.

Within open-source models, the medium group shows the clearest stratification. Qwen3-VL-8B presents the most uniformly high profiles in both question and video consistency, whereas VideoChatGPT and Phi-4-Multimodal display consistently low and uneven shapes, suggesting that improvements in standard QA accuracy do not necessarily translate into stable contrastive decision patterns. The small-model group further highlights this mismatch: despite moderate BaAcc. For commercial models, the group shows substantially more balanced radar shapes in consistency metrics, with GPT-5-mini and Gemini-2.5 variants maintaining stronger and more uniform performance across categories, while GPT-5-nano remains weaker and less stable, especially on the harder Counterfactual and Causal axes.

\begin{figure}[h!]
  \centering
    \centering
    \includegraphics[width=\linewidth]{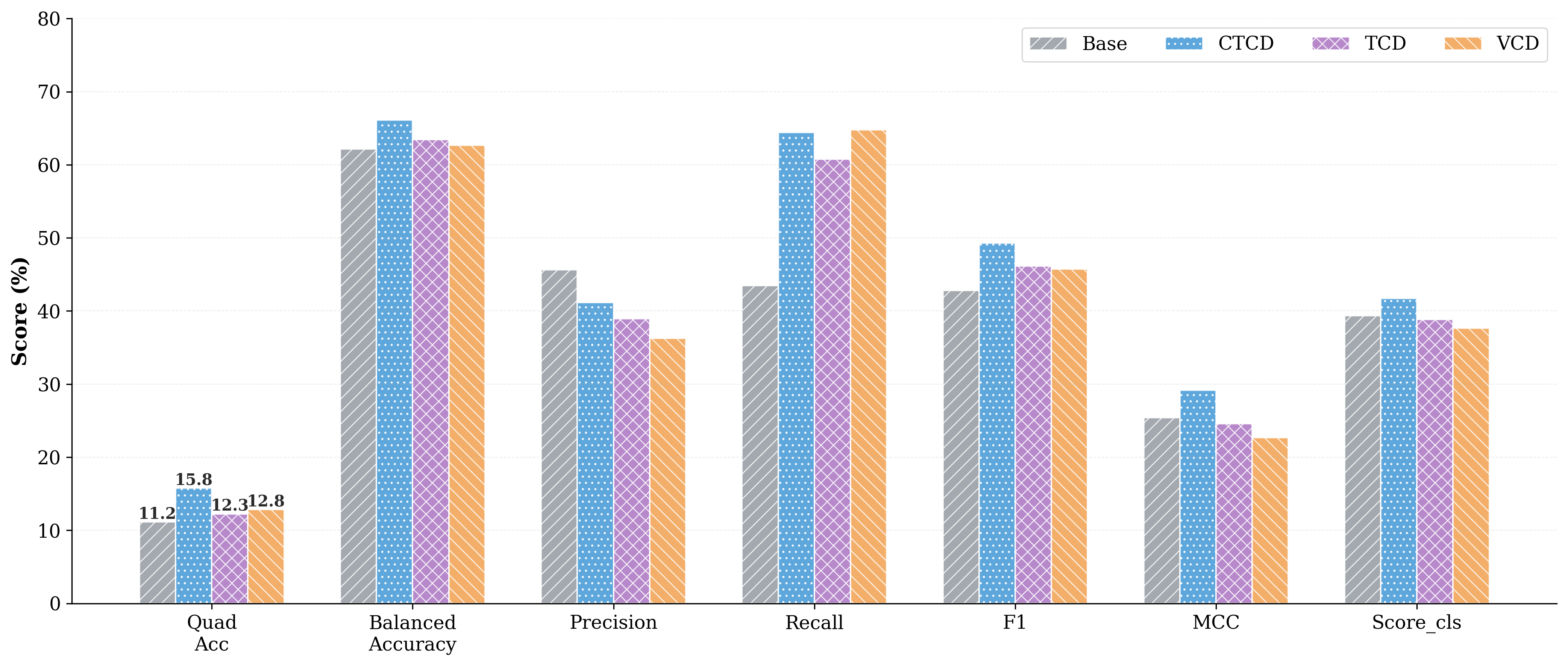}
    \caption{Qwen3-VL contrastive decoding comparison.}
    \label{fig:cd-qwen}
  \hfill

  \caption{Grouped bar charts for Qwen3-VL-7B  contrastive decoding methods, Quad Acc and classification metrics.}
  \label{fig:cd-methods-grouped}
\end{figure}

\begin{figure*}[h!]
  \centering
  \includegraphics[width=\textwidth]{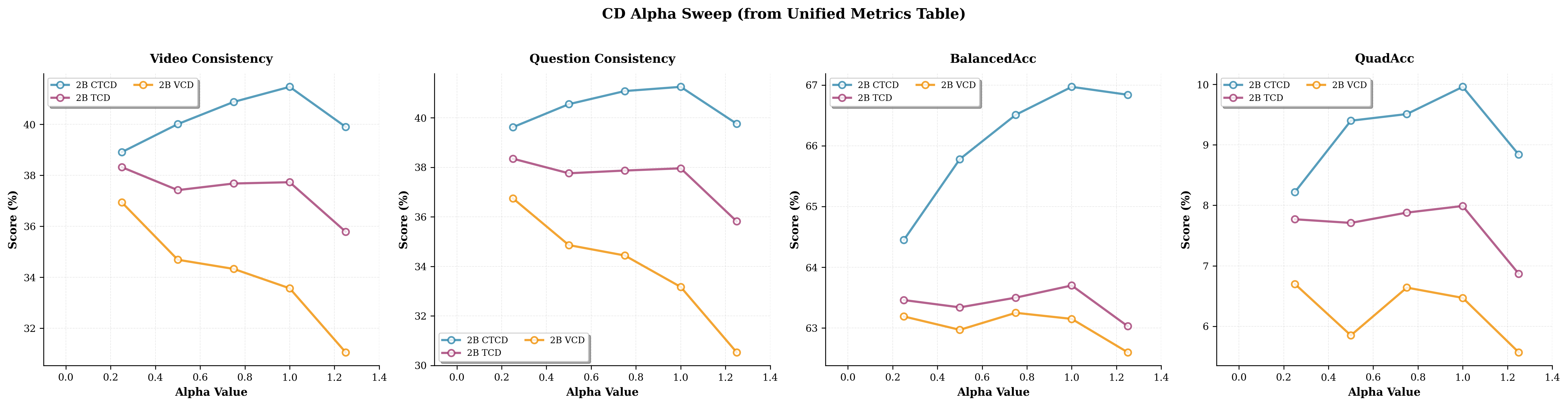}
 \caption{Effect of contrast strength $\alpha$ on Qwen3-VL-2B under VCD, TCD, and C-TCD. We report video consistency, question consistency, balanced accuracy, and QuadAcc.}
 \label{fig:cd_classification_alpha_tuning}
\end{figure*}


\subsubsection{Detailed Contrastive Decoding Results}
Fig.~\ref{fig:cd-qwen} reveal that contrastive decoding improves standard QA scores but is not always aligned with quadruple consistency. On Qwen, C-TCD yields the strongest overall gains, increasing QuadAcc while also improving balanced accuracy, F1, MCC, and Score$_{\text{cls}}$. The main effect is a large boost in recall, which indicates that Qwen is partially omission-limited on positive scenes and benefits from a semantically decisive contrast signal provided by the paired counterfactual video. In comparison, VCD and TCD also raise recall but reduce precision more noticeably, suggesting that generic corruption or temporal degradation can over-suppress useful evidence and shift the decision boundary in a less targeted way than C-TCD.

\subsubsection{Hyperparameter Analysis of Contrastive Decoding}

Fig.~\ref{fig:cd_classification_alpha_tuning} analyzes the effect of the contrast strength $\alpha$ for Qwen3-VL-2B under VCD, TCD, and C-TCD. Overall, C-TCD is the most $\alpha$-sensitive but also the most rewarding: increasing $\alpha$ steadily improves both video consistency and question consistency, and these gains translate to higher BalancedAcc and QuadAcc up to a clear sweet spot around $\alpha\approx 0.75\text{--}1.0$. Beyond this range, performance drops across metrics, suggesting that overly strong contrast begins to over-suppress valid evidence and destabilizes the strict quadruple decision pattern. TCD is comparatively stable across $\alpha$ with modest fluctuations, indicating that temporal degradation provides a weaker and less targeted contrast signal, so scaling $\alpha$ yields limited returns. In contrast, VCD degrades as $\alpha$ increases, with both consistency scores and QuadAcc trending downward, which is consistent with visually corrupted negatives becoming increasingly harmful when their influence is amplified. These results support using a moderate contrast strength for reliable improvements, and they further highlight that semantically decisive counterfactual pairs make C-TCD substantially more effective and tunable than generic corruption or temporal degradation.

\subsection{Qualitative Results}

\begin{figure*}[ht]
  \centering
  \includegraphics[width=\textwidth]{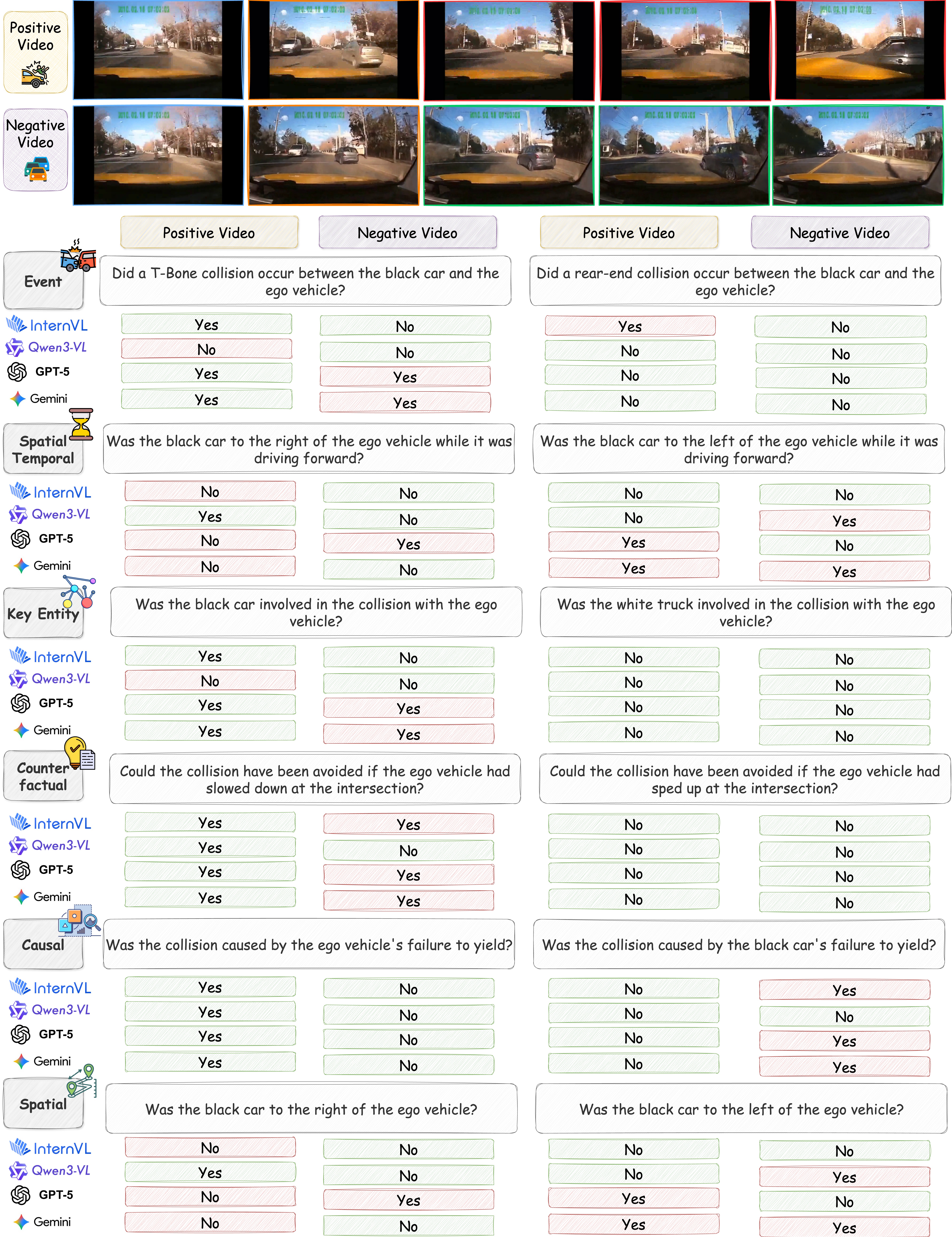}
  \caption{Qualitative example 1: Intersection T-bone collision between the black car and the ego vehicle. }
  \label{fig:qual_case1}

\end{figure*}

Fig.~\ref{fig:qual_case1} shows an intersection scenario with a T-bone collision between the black car and the ego vehicle in the positive video, paired with a counterfactual negative where the collision is removed. The quadruple structure exposes two common inconsistencies: some models correctly fire on the target event but still answer \emph{Yes} on the counterfactual negative under the same query, while others confuse the mutually exclusive alternative collision type and trigger a false positive on the rear-end hypothesis on the same scene. The spatial and spatial-temporal pairs further reveal brittle left–right and front–behind grounding under near-identical views, where models may change answers across videos but in the wrong direction.

\begin{figure*}[h]
  \centering
  \includegraphics[width=\textwidth]{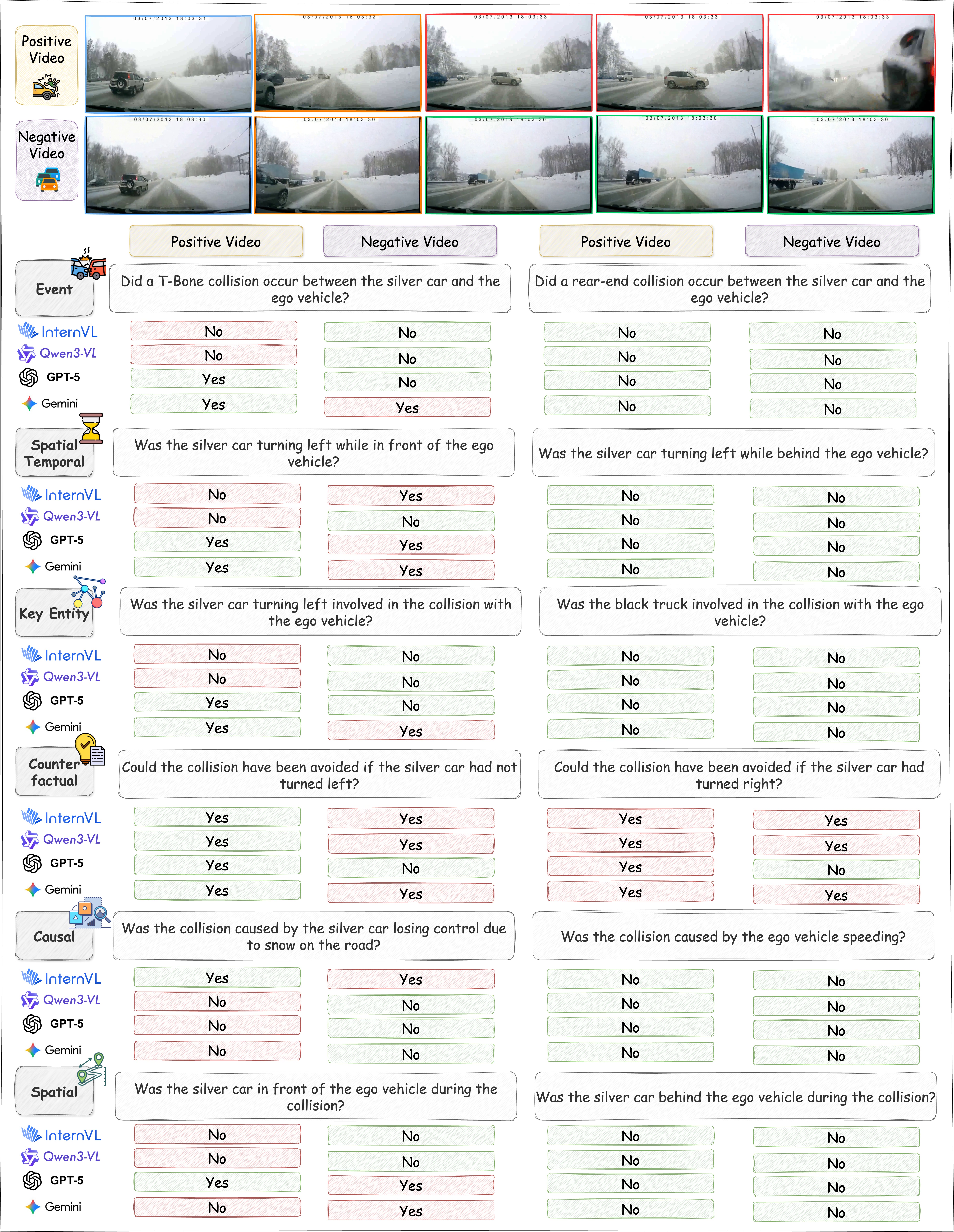}
  \caption{Qualitative example 2: Snowy-road turning scenario with loss of control collision.}
  \label{fig:qual_case2}
\end{figure*}

Fig.~\ref{fig:qual_case2} presents a snowy road case where the silver car’s turning maneuver is central to the accident trajectory. This scene highlights how temporal-localized motion cues and turning intent can dominate the downstream causal and counterfactual questions: even when models reject the alternative event type, they may still fail the none-of-the-above requirement by producing affirmative answers on the counterfactual negative for maneuver-dependent or explanation-style queries. The case illustrates that counterfactual and causal questions are especially sensitive to subtle premise mismatches and to ambiguity in what visual evidence is sufficient to justify a causal attribution.

\begin{figure*}[h]
  \centering
  \includegraphics[width=\textwidth]{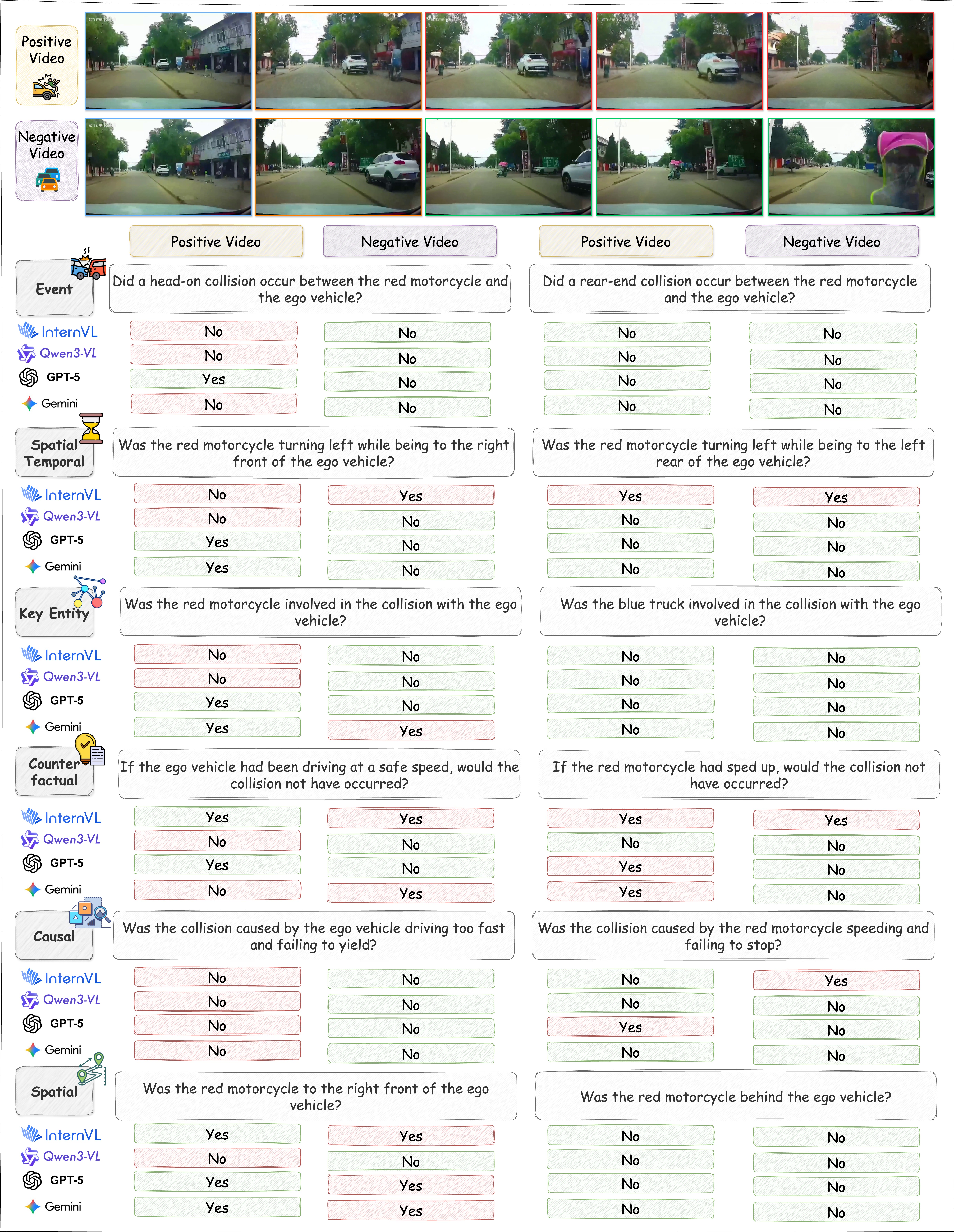}
  \caption{Qualitative example 3: Urban scenario with a head-on collision between a red motorcycle and ego vehicle.}
  \label{fig:qual_case3}
\end{figure*}

Fig.~\ref{fig:qual_case3} shows an urban scene involving a red motorcycle, where the target event is a head-on collision in the positive video and the paired negative removes the collision while preserving the surrounding layout. Here, a frequent failure mode is \emph{omission} on the positive video, where models default to \emph{No} for the target event and propagate that conservatism to related key-entity and counterfactual questions. At the same time, some models still hallucinate on the counterfactual negative for counterfactual-condition queries, suggesting that plausible textual hypotheses can override the intended none-of-the-above state even when the paired video edit removes the critical event.

\end{document}